\documentclass{article}

\usepackage{microtype}
\usepackage{graphicx}
\usepackage{subfigure}
\usepackage{array}

\usepackage{booktabs} 

\usepackage{hyperref}

\usepackage[accepted]{icml2025}

\usepackage{amsmath}
\usepackage{amssymb}

\usepackage{mathtools}
\usepackage{amsthm}

\usepackage[capitalize,noabbrev]{cleveref}

\theoremstyle{plain}

\theoremstyle{definition}

\theoremstyle{remark}

\usepackage[textsize=tiny]{todonotes}

\begin{document}

\twocolumn[
\icmltitle{Develop AI Agents for System Engineering in \textit{Factorio}}



\icmlsetsymbol{equal}{*}

\begin{icmlauthorlist}
\icmlauthor{Neel Kant}{hai}
\end{icmlauthorlist}

\icmlaffiliation{hai}{Hippocratic AI}

\icmlcorrespondingauthor{Neel Kant}{neel@hippocraticai.com}

\icmlkeywords{Machine Learning, Agentic AI, Systems Engineering, Reinforcement Learning, Adaptability, Open-Ended Simulations, Complex Systems, Industrial Automation, AI Infrastructure, Factorio, Continuous Learning.}

\vskip 0.3in
]



\printAffiliationsAndNotice{}  

\begin{abstract}
Continuing advances in frontier model research are paving the way for widespread deployment of AI agents. Meanwhile, global interest in building large, complex systems in software, manufacturing, energy and logistics has never been greater. Although AI-driven system engineering holds tremendous promise, the static benchmarks dominating agent evaluations today fail to capture the crucial skills required for implementing dynamic systems, such as managing uncertain trade-offs and ensuring proactive adaptability. \textbf{This position paper advocates for training and evaluating AI agents’ system engineering abilities through automation-oriented sandbox games—particularly \textit{Factorio}.} By directing research efforts in this direction, we can equip AI agents with the specialized reasoning and long-horizon planning necessary to design, maintain, and optimize tomorrow’s most demanding engineering projects.
\end{abstract}

\section{Introduction}
Since the release of ChatGPT in November 2022, the field of generative AI has experienced an explosive surge in both attention and investment. Early successes in large language models (LLMs) have demonstrated that tuning foundation models to follow instructions and optimize for human preferences can yield AI systems capable of a wide range of tasks—often approaching or matching human-level proficiency in specific domains. This has led researchers and industry experts alike to speculate that we now possess the fundamental building blocks for artificial general intelligence (AGI).

A logical progression beyond chatbots and prompt-based LLMs is the development of \textbf{AI agents}. Unlike conventional models that simply output text in response to queries, agentic AI systems combine language comprehension with memory, tools, and other interfaces, allowing them to interact with environments in near-human ways. This paradigm shift has ignited excitement about the possibility of autonomous, always-on AI “workers” that can undertake many tasks currently performed by humans—ranging from data analysis to coding, from supply-chain management to design optimization.

Yet, for all the excitement around AI agents, today’s systems often remain limited in scope and capability. Part of this is due to a lack of robustness and a need for continued integration with real-world interfaces, but it is also due to limitations of static development environments for agents. We argue that these constraints can be overcome by explicitly training and evaluating agents in system engineering tasks, where scalability, adaptability, and long-term strategic thinking become paramount. By building, optimizing, and maintaining complex, real-world systems—or close simulations thereof—agents can push well beyond static benchmarks toward generalized \textbf{superhuman problem-solving}.

This paper makes three main points:

\begin{enumerate}
    \item \textbf{System engineering} is a uniquely high-leverage capability. Societies worldwide face challenges that demand new levels of coordination and innovation in designing and managing complex infrastructures and processes.

    \item \textbf{Sandbox-style simulation platforms} are essential for training and evaluating AI agents on their capacity to handle real-world complexities. Such platforms can capture the interplay between \emph{adaptability}, \emph{automation}, and other \emph{dynamic trade-offs} that static benchmarks fail to represent, thereby enabling more realistic and robust testing.

    \item \textbf{\textit{Factorio}} stands out as the ideal sandbox game for this purpose as its entire nature centers on designing and automating complex systems along with key technical advantages like robust support for modifying and augmenting game mechanics.
\end{enumerate}

We explore each of these points in detail in subsequent sections individually, and provide an Appendix which visually illustrates \textit{Factorio} for newcomers to gain intuition about its gameplay.
\section{The Importance of System Engineering}

In this section, we examine trends in complex system development and AI agents. We deduce that these will converge and we thus firmly establish the value proposition for developing system engineering capability in AI agents.

\subsection{The Ubiquity of Systems}

A system is defined as a collection of interacting components that together serve a function or purpose. Under this broad definition, the world as we know it is held together by complex systems. Key examples include:
\begin{itemize}
    \item \textbf{Transportation and Logistics.} Industries such as shipping, trucking, and ridesharing; public services like trains and buses; physical infrastructure projects including roads, bridges, and tunnels.
    \item \textbf{Energy Infrastructure.} Raw material acquisition and refinement; large-scale energy generation in specialized facilities; storage and distribution networks required for load balancing.
    \item \textbf{Modern Agriculture.} Encompassing land management (irrigation, fertilization, pest control), crop and livestock cycles, as well as packaging and distribution systems to bring products to market.
    \item \textbf{Advanced Manufacturing.} The creation of parts from raw materials; international supply chain coordination; final assembly of complex products across sectors such as computing, biotechnology, and aerospace.
    \item \textbf{Digital Ecosystems.} Physical networking infrastructure; internet hosting servers; cloud computing stacks; software frameworks, libraries, and algorithms.
\end{itemize}

\textbf{System engineering} refers to the design, implementation, and management of such large-scale systems that integrate hardware, software, and human processes. Our understanding of systems has evolved significantly over time. Early industrial breakthroughs (e.g., standardized components, assembly lines, electrification) introduced new layers of complexity by increasing production volumes, lowering costs, and extending distribution chains. In aerospace and defense projects, where massive interdisciplinary teams had to be coordinated, formal \textit{system engineering} practices emerged \cite{blanchard2010, kossiakoff2011, incose2015, buede2016, madni2018}. Over the years, these yielded iterative and agile methods emphasizing continuous integration and rapid feedback—trends that are now mainstream in digital ecosystems. 

The demand for building and scaling complex systems shows no signs of slowing down. Across the world, large projects are planned or underway to address unprecedented challenges in the form of energy needs, aging demographics, changing climate patterns, geopolitical tensions, and more \cite{mckinsey2022, deloitte2023, reshoring2023}. The competitive advancement of technology itself leads to self-reinforcing demand for systems, exemplified by staggering investments into AI-related infrastructure \cite{uscongress2022, stargate2025}. Budget overruns and timeline delays are all too common in implementing large projects, showing that human planning and system engineering has its limitations. It thus appears nearly certain that advanced AI will be crucial in tackling these challenges and implementing the solutions.

\subsection{The Rise of AI Agents}

Simultaneously, AI agents are gaining traction as the most promising framework for applying human-aligned generative AI models. Significant advances in multimodal input processing and reasoning through inference-time compute use have exposed the possibilities of autonomous agents using complex interfaces to accomplish tasks over longer time horizons. These opportunities are rapidly being realized through virtual agents which are increasingly using web browsers, code interpreters and other digital tools to automate workflows and offload other labor from humans \cite{devin2024, openai2025operator}. Progress in physical agents is also picking up steam, leaning on advances in both general-purpose foundation models and maturing robotics technology \cite{nvidia2024gr00t, musk2024optimus}. 

The tasks assigned to AI agents today may be composites of several smaller sub-tasks, but ultimately tend to be self-contained workflows. As AI agents become more reliable at executing these tasks, it will become more enticing to involve them in more system-level challenges. Systems-level expertise is always more scarce since it demands deep knowledge of choices for components, interconnections and associated trade-offs for costs, implementation time, complexity, scalability, etc. Training data for reasoning about systems is also commensurately scarce and so it will naturally present a challenge for improving AI agents. As AI agents proliferate, the challenges and opportunities associated with multi-agent coordination will also become more relevant and influence the efficacy of AI-enhanced systems. Hence, this trajectory of elevating AI agents to effectively work on system engineering seems central for achieving the long-term goals of developing AGI. 

\subsection{Implications of Superhuman System Engineering}

AI models and agents have shown superhuman ability in various domains. Historically, this is evidenced by their mastery of classically challenging board games and more complex real-time strategy video games \cite{silver2017alphazero, vinyals2019alphastar}. In applied settings, AI models have outperformed human-engineered solutions in predicting molecular structures \cite{alphafold2021}, weather patterns \cite{gencast2024}, and even designing certain GPU circuits \cite{prefixrl2021}. Recent frontier models demonstrate advances in multimodal and highly technical reasoning, suggesting that a trajectory toward general superhuman intelligence is potentially close. Given all this, it is worth considering the ramifications of successfully building superhuman AI system engineers.

We can look to current examples of top-tier system engineering by humans. These achievements share a common pattern: they redefined what was previously considered possible. For instance, when Apple replaced Intel chips with its in-house M-series CPUs, the gains in thermal performance, battery life, and software speed were widely seen as a generational leap. Similarly, SpaceX reshaped the frontier of aerospace by reducing costs by orders of magnitude and inventing reusable rocket technology. Meanwhile, the unrelenting dominance of Nvidia’s entire data center stack—from hardware to deep learning libraries—has made it (at least temporarily) the most valuable company in the world. In each of these cases, the teams in charge took ownership of the entire system, jointly optimizing it over many iteration cycles to achieve superior metrics. It is worth stressing that the same depth of expertise required for proposing a comprehensive initial design is needed for continually refactoring a system to improve its scalability, maintainability, security, and fault tolerance as development progresses

Superhuman AI agent capability in system engineering gives us a much better chance of addressing civilizational challenges, such as scaling clean energy systems, securing reliable water and food supplies, and lowering the cost of economically important finished goods. In other words, superhuman-level system engineering is the key to producing utopian abundance (provided we solve the alignment problem for such superintelligence). Full automation of physical projects will also require general-purpose robotics, which may be a bottleneck in the near term. Yet as those technologies mature, AI agents could combine high-level system design with low-level mechanical tasks, delivering a fully optimized, end-to-end engineering capability for arbitrarily complex systems. 
\section{Designing Evaluations for System Engineering}
We highlight core trade-offs associated with building systems, namely efficiency, scalability and adaptability. We argue that system engineering training and evaluation environments must be \textit{dynamic} and \textit{open-ended} to adequately assess the dynamic equilibrium of these characteristics.

\subsection{Real-World Intuition}
In the design phase of a system engineering project, the focus is on delivering a proposal that meets various requirements and user preferences for features and costs. This requires deep domain expertise since many valid proposals can exist, yet vary in terms of up-front costs, maintenance costs, implementation time, complexity, regulatory compliance, scalability, and so on. Design capability is readily tested in the software industry with system design interviews that pose questions such as “How would you design a real-time collaborative word processing application like Google Docs?” or, more bluntly, “Design Google Docs.”, “Design Uber.”, “Design Twitter.”, etc. These questions are not meant to be answered in a single pass, but rather serve as a starting point for iteratively gathering requirements and proposing increasingly detailed solutions.

\begin{figure}[ht]
    \centering
    \includegraphics[width=\columnwidth]{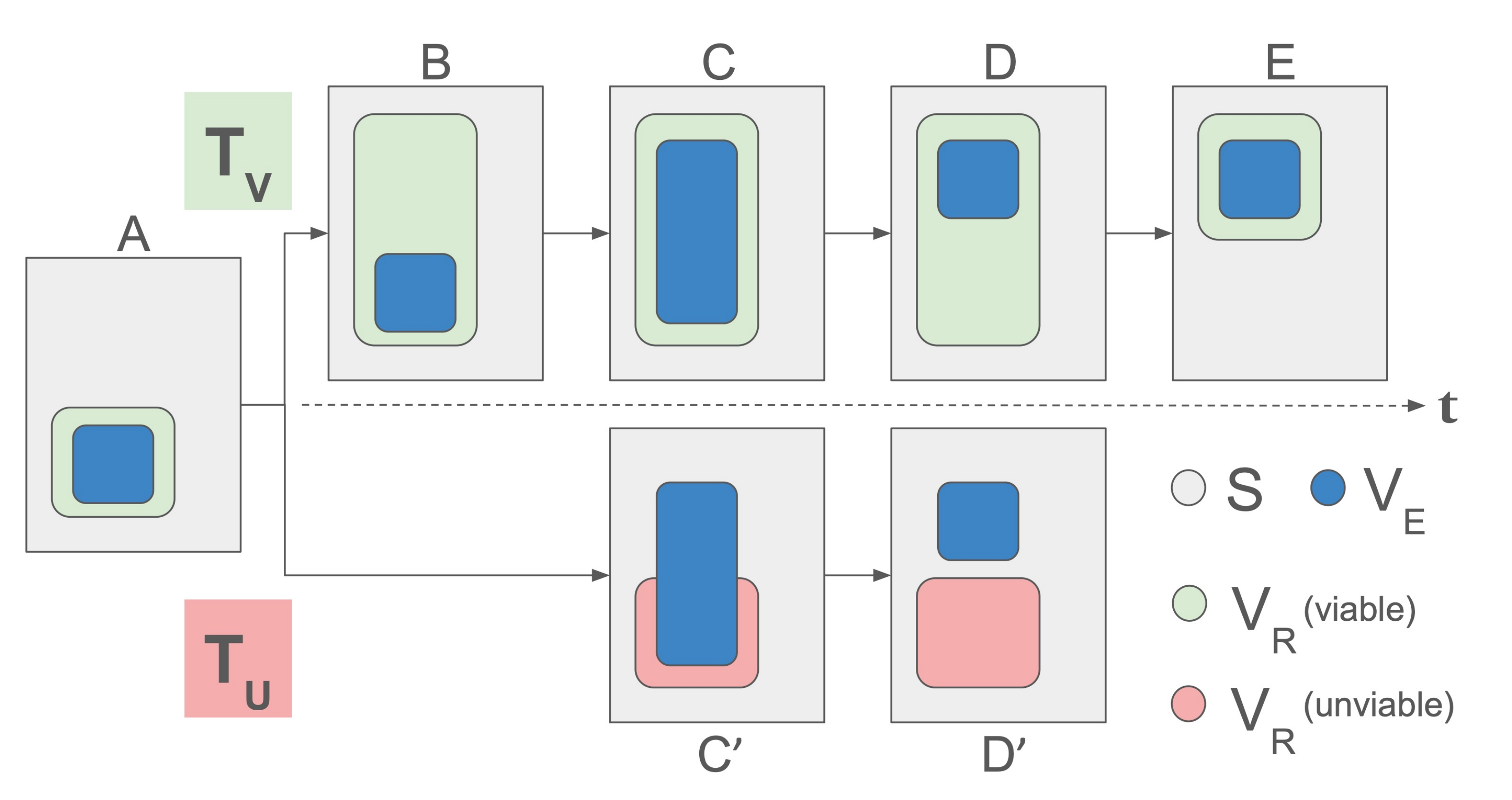}
    \caption{\textit{The Law of Requisite Variety.} $T_V: A \rightarrow E$ is a trajectory where a system stays \textbf{viable} through adaptation. $T_U: A \rightarrow D\prime$ shows an alternate trajectory where the system does not adapt and becomes \textbf{unviable}. A system is viable within the total state space $S$ when the variety of the environment at that time $V_E$ remains a subset of variety the system can handle $V_R$. Systems must adapt proactively ($A \rightarrow B$) to ensure this condition is met, but then ideally reduce variety to improve efficiency and maintainability ($D\rightarrow E$).}
    \label{fig:requisite_variety}
\end{figure}

After a real system is designed, the implementation phase begins and often never truly ends. Successful systems typically continue to expand in scope because increased outputs fuel greater demand. This pattern is evident in large software services, energy networks, and public transportation systems. Even if overall scale plateaus, there is an ongoing need for repair and maintenance—particularly in physical systems but also in software, which must periodically upgrade dependencies and refactor for performance. Consequently, the longevity and effectiveness of a system fundamentally depend on its capacity to assimilate feedback and \textit{adapt} to inevitable changes.

Feedback collection is facilitated through automated means like logging in software or more manually such as accepting verbal customer feedback. Adapting the system with this feedback is thus core to ensuring it meets expectations through key performance indicators. Some future scenarios are more serious and difficult to fully predict. Recent examples such as the COVID-19 pandemic required large-scale adaptations not seen since World War II, and the volatility of geopolitics—as highlighted by the conflict in Ukraine—continues to demand swift adjustments in global systems. Natural disasters like hurricanes and wildfires, technological breakthroughs such as the generative AI boom, major cybersecurity incidents, and new discoveries of key commodities further underscore the need for flexible system design.

\subsection{Supporting Theory}
Fortunately, the study of systems has long acknowledged the value of adaptability, leading to foundational frameworks that inform real-world solutions. One such lineage is \emph{cybernetics} \cite{wiener1948}, which reveals how continuous feedback loops and robust communication channels allow systems to counter external disturbances. Ashby’s \textbf{law of requisite variety (LRV)} \cite{ashby1956} stresses that systems must possess enough complexity (known as \textit{variety}) internally to handle the complexity of potential external disruptions (Figure \ref{fig:requisite_variety}. If this condition is not met, it can lead to a loss of \textit{stability} of the system, meaning that it will not be able to maintain desired indicators of success. The intuition is comparable to that of machine learning theory, where out-of-distribution inputs lead to poor model performance. 

\begin{figure}[ht]
    \centering
    \includegraphics[width=\columnwidth]{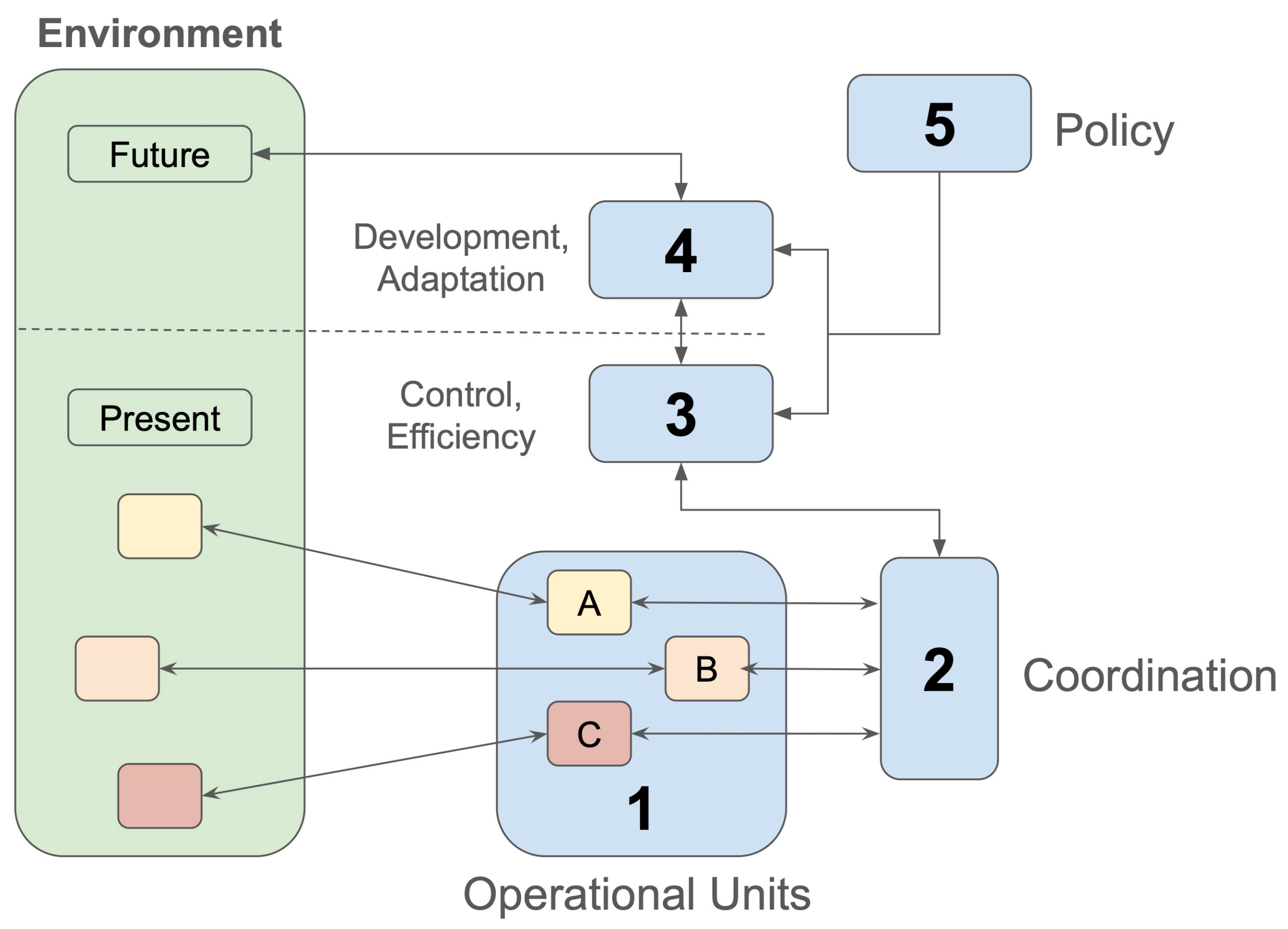}
    \caption{\textit{The Viable System Model.} Systems are organized into five levels concisely given as: \textit{1. operational units, 2. coordination, 3. present optimization, 4. future planning, and 5. ultimate policy.} \textbf{See Table~\ref{table:vsm_table}} for longer descriptions. These levels are only responsible for the \textit{variety} associated with that level and can escalate or delegate as needed. A key aspect is how Level 5 effectively balances out the tension between Levels 3 and 4 which are more present- and future-focused respectively.}
    \label{fig:viable_system}
\end{figure}

The law is typically presented in the static setting, meaning it applies to the (internal) response variety ($V_R$) and environmental variety ($V_E$) at any given time. However, it can be extended to apply over time, where the configuration of a system must be able to change in order to support the particular variety of the environment over time (Figure~\ref{fig:requisite_variety}). Building on this, Beer's \textbf{viable system model (VSM)} \cite{beer1959, beer1972} emphasizes hierarchical structures for robust systems. The modularity of hierarchy allows different levels of a system to handle only the variety of inputs which the level is responsible for (Figure \ref{fig:viable_system}, Table~\ref{table:vsm_table}). For example, the lowest level (System 1) of viable systems are the autonomous operational units which act in the world, so they individually only need to support their distinct low-level functions. In this model, it is essential for systems to have a layer which plans \textit{proactive} \textit{adaptation} (System 4), enabling organizations and infrastructures to pivot swiftly under changing requirements. 

The law of requisite variety (LRV) and the viable system model (VSM) highlight a central tension in robust system operation: \textit{efficiency} and \textit{flexibility} tend to come at the cost of each other. For instance, a mechanized assembly line can mass-produce a single product more rapidly than a human worker, yet the latter may be more versatile in producing a variety of items. In software, production-level code is often streamlined through rigid abstractions, whereas one-off scripts are less optimized but highly flexible. Even in the study of LLMs, the choice between prompt-engineering large models and finetuning smaller ones reflect this same trade-off. This principle is shown graphically in Figure~\ref{fig:requisite_variety} where system variety $V_R$ is expensive to maintain, and ultimately should be reduced when unneeded. Likewise, Figure~\ref{fig:viable_system} illustrates that System 3 and 4 directly embody this tension and it is up to System 5 to arbitrate and maintain cohesion. Scaling up and maintaining systems thus presents a persistent challenge of preserving \emph{dynamic equilibrium}, in which the benefits of automation and scale do not compromise a system’s capacity to adapt \cite{forrester1961industrial, holling1973resilience, sterman2000business}.

From a machine learning perspective, adaptability has been explored under many paradigms, including domain adaptation \cite{redko2022domainadaptationtheory}, meta-learning for agents \cite{beck2024metareinforcementlearning}, continual learning \cite{wang2024continuallearning}, in-context learning \cite{dong2024incontextlearning}, and out-of-distribution generalization \cite{liu2023outofdistribution}. Central themes across these fields involve developing robust representations, ensuring sample-efficient training, and promoting safe exploration. By weaving AI-driven automation into systems, we now have the opportunity to significantly enhance both efficiency and adaptability—two objectives that have traditionally been at odds.

\subsection{Evaluations for AI Agents}
Recent advances in AI research have fueled efforts to build virtual agents capable of increasingly complex interactions with real-world interfaces. As these cognitive capabilities continue to mature, they provide a foundation for agents to meaningfully contribute to system engineering projects. Realizing this vision, however, requires a fundamental rethinking of how we both train and evaluate virtual AI agents.

Evaluation methods for LLM-derived agents naturally began with classic NLP benchmarks, such as question answering in MMLU \cite{hendrycks2021mmlu}. They have since evolved to encompass multi-turn interaction \cite{zheng2023mtbench}, multimodality \cite{yue2024mmmu}, and external tool use \cite{zhou2023agentbench, he2024webvoyager}—capabilities expected of advanced AI agents. SWE-bench \cite{jimenez2024swebench} (and its multimodal extension \cite{yang2024swebenchmultimodalaisystems}) is likely the most challenging agent benchmark in use today. It requires agents to resolve issues in codebases by modifying multiple files and subsequently passing unit tests. Though it involves reasoning and multi-step planning, it remains a static evaluation that does not measure the capacity to maintain dynamic equilibrium between VSM Systems 3 and 4 and deal with the uncertainty of dynamic environment variety as per the LRV. This would hold true even for an extension of the benchmark in which agents designed a system like Google Docs and implemented it, yet never had to respond to changing requirements or circumstances.

By contrast, non-LLM-based agents have often been evaluated in \emph{dynamic} environments. This is the case for agents achieving superhuman performance in competitive games such as Go \cite{silver2017alphazero} and StarCraft II \cite{vinyals2019alphastar}, where the presence of an opponent forces rapid adaptations to both the agent’s own actions and those of adversaries. For a time, increasingly complex games appeared to be a promising route to building general intelligence, culminating in work on \textit{Minecraft} via Voyager \cite{wang2023voyager} and MineDojo \cite{fan2022minedojo}. These agents achieved goals in a \emph{dynamic}, \emph{open-ended} environment, with effectively unconstrained objectives demanding resource gathering, multi-step planning, and adaptability to emergent challenges. 

\subsection{The Ideal Evaluation Environment for System Engineering}
Interest in dynamic, open-ended environments waned somewhat after the advent of LLM-based generalist models. However, the rapid evolution of ChatGPT and its successors—featuring multimodality, tool-use capabilities, and ample test-time compute—opens new possibilities for resurrecting this research agenda in a more advanced form.

We deduce from the ar, \emph{sandbox games} which support \textit{automation} as a mechanic are the ideal setting for evaluating system engineering. They let researchers specify high-level objectives and observe an agent’s ability to break down tasks, weigh trade-offs, and implement solutions. Over time, the researcher can change these objectives or introduce disruptions, testing the agent’s capacity to maintain a healthy dynamic equilibrium as per the viable system model. Greater open-endedness is also desirable as it allows for more comprehensive testing of an agent's ability to comply with the law of requisite variety. Simulated environments additionally have the benefits of being fundamentally safer than real-world testing and can manage the trade-off between world physics complexity and scalability.

{
\hyphenpenalty=10000
\exhyphenpenalty=10000
\begin{table*}[htbp!]
\centering
\small
\begin{tabular}{l|p{4cm}p{8.5cm}}
\hline

\noalign{\vskip 2pt}
\textbf{VSM Level} & \textbf{Responsibility} & \textbf{Factorio Example} \\
\noalign{\vskip 2pt}
\hline
\noalign{\vskip 1pt}
\textbf{System 1} &
Front-line operations; directly transform inputs into outputs &
Assemblers, miners, and furnaces that convert raw materials (e.g.\ iron ore) into plates and intermediate products. These are the basic production units forming the backbone of the factory. \\
\noalign{\vskip 1pt}
\hline
\noalign{\vskip 1pt}
\textbf{System 2} &
Coordinates and stabilizes System 1 units &
Conveyor belts, splitters, and simpler logistic setups to route materials between different production areas, prevent bottlenecks, and ensure each assembler or furnace receives the resources it needs. \\
\noalign{\vskip 1pt}
\hline
\noalign{\vskip 1pt}
\textbf{System 3} &
Manages and allocates resources, drives efficiency, ensures smooth operation &
Monitoring production levels, adjusting supply lines to balance throughput, and deploying construction/logistics bots for on-demand tasks such as repairs or setting up new sections. This maintains overall operational stability. \\
\noalign{\vskip 1pt}
\hline
\noalign{\vskip 1pt}
\textbf{System 4} &
Plans expansions, researches new technology, foresees future needs &
Choosing research paths (e.g.\ robotics, nuclear power), planning additional outposts for resource gathering, and redesigning factory layouts to handle increased demand or optimize long-term efficiency. \\
\noalign{\vskip 1pt}
\hline
\noalign{\vskip 1pt}
\textbf{System 5} &
Sets overall purpose, policy, and alignment &
Defining the ultimate mission (e.g.\ launching a rocket by a target time), deciding on environmental constraints (such as minimizing pollution), and determining the overarching strategy (e.g.\ peaceful or militaristic). \\
\noalign{\vskip 1pt}
\hline
\end{tabular}
\caption{Viable System Model (VSM) levels mapped to Factorio examples.}
\label{table:vsm_table}
\end{table*}
}

Drawing on \textit{Minecraft} as inspiration, one can envision an “ideal” environment that focuses on abstractions relevant to system engineering while omitting excessively detailed physics. Full 3D simulations can be computationally expensive and often distract from the higher-level reasoning crucial for scaling and process orchestration. Accordingly, a game environment centered on resource flows, balancing trade-offs, and long-horizon planning is preferable. Core properties of such an environment include:
\begin{itemize}
    \item \textbf{Automation.} The agent’s action space should permit automating processes and managing the associated trade-offs between efficiency and adaptability. This is key for testing System 3 and 4 capability as per the VSM.
    \item \textbf{Complex Evaluation Metrics.} Long-horizon performance, resource usage, and resilience under partial failures become measurable, enabling richer assessments than single-turn tests. This is part of high environment variety in the LRV.
    \item \textbf{Multi-Agent Support.} Collaboration with peers, hierarchical coordination, and competition with adversaries significantly increase complexity, further testing an agent’s capacity to adapt. This is also key for testing System 3 and 4 capability in the VSM. 
    \item \textbf{Modding Support.} Allowing users and artificial agents to create modifications or extensions fosters adaptation to out-of-distribution scenarios. This another way to have high environment variety in the LRV.
    \item \textbf{Scalability.} The environment mechanics should be at the right level of abstraction to facilitate systems reasoning, planning, and implementation without requiring excessive computational resources.
\end{itemize}

There are many candidate sandbox games—\textit{Cities: Skylines}, \textit{The Sims}, \textit{Stardew Valley}, \textit{Kerbal Space Program}, \textit{No Man's Sky}, \textit{Satisfactory}, among others—that support a form of system engineering. Yet they each have limitations with respect to one or more of the above criteria. As the next section will show, \emph{Factorio} stands out for providing an ideal testbed for AI system engineering: its mechanics inherently encourage large-scale “megabase” building, resource management, automation, and iterative adaptation.

\section{Factorio as a System Engineering Testbed}
We now argue that \textit{Factorio} is an ideal environment to develop system engineering capability in AI agents. We describe the mechanics, features and extensible scope of the game and put forth a call for using \textit{Factorio} as a platform for public research. Interested readers can find a more detailed walkthrough of the game in Appendix~\ref{ref:sec_appendix}

\subsection{Overview}
\textit{Factorio} is a 2D, top-down factory-building game that centers on \textit{automation}, rendering it a uniquely rich environment for developing and evaluating AI agents with strong system engineering capabilities. Although it shares the open-sandbox approach of titles like \textit{Minecraft}, \textit{Factorio} is far better suited for this purpose as it emphasizes building systems with high throughput, efficiency, and resilience. Automating the production of goods—from early hand-assembled items to complex industrial chains—is not merely a side option but rather the heart of the gameplay. This emphasis on scaling and optimizing factories pushes agents to navigate challenges that mirror real-world engineering dilemmas: resource constraints, energy usage, logistical complexity, and even defensive measures against hostile forces.

A key metric of success is \textbf{science per minute} (SPM), a community-standard indicator of a factory’s overall efficiency in generating the science packs needed for technological progress. Because each successive tier of research unlocks new possibilities (e.g., improved assemblers, trains, robots) but also imposes heavier resource and energy demands, any small inefficiency can ripple into crippling bottlenecks. Consequently, an effective agent must maintain the appropriate degree of variety in its approach at all times, ensuring that its decision-making processes can handle the game’s growing complexity and unexpected fluctuations. SPM makes for a great summary benchmark metric, with human novice bases at $\sim$0-30 SPM, intermediate at $\sim$30-200 SPM and advanced bases at $\sim$200-1000+ SPM. 

From a VSM perspective, \textit{Factorio} initially starts players at purely System 1 activities like manually extracting coal and iron to be placed in a hand-crafted furnace. The use of automated conveyer belts with splitting and load-balancing mechanisms combined with automated inserter arms elevates design to a System 2 level. A key gameplay entity is the \textit{assembler} which can be programmed with a recipe to convert inputs to finished outputs using materials, power and space for operation. Scaling the automated production of intermediate goods setting up train cargo networks and selecting technology tree paths are all System 3, 4 and 5 functions (Table~\ref{table:vsm_table}). A pivotal late game technology is the use of \textit{automated construction robots} which can be used to rapidly bring and place materials in accordance with large, complex player-made blueprints. This capability hence focuses gameplay purely on systems-level control problems, choosing the right smelting column, railway depot, solar array configuration, etc. to evolve the base as needed.

This versatility is further magnified by \textit{Factorio}’s robust modding support, which allows researchers and the broader community to introduce new mechanics, custom APIs, or entire rebalanced rule sets. In other words, the sandbox nature of \textit{Factorio} can be extended indefinitely, enabling the environment itself to evolve and stress-test an agent’s capacity to adapt and manage variety. Such flexibility in scaling and customization makes \textit{Factorio} ideal for public research, as it encourages the development of AI agents that can grow beyond initial, narrowly-defined tasks and rise to dynamic challenges that demand integrated System 1–5 competencies.

\subsection{Challenges for Current AI Agents}

While AI agents have made remarkable progress in reasoning and multimodal interaction, there remains a sizable gap between the capabilities of frontier agents (e.g. \cite{deepmind2024mariner, openai2025operator}) and the level of sophistication needed to thrive in \textit{Factorio}—and, by extension, in complex real-world systems. For example, \textit{Factorio} uses traditional a keyboard-and-mouse interface with numerous GUI windows and features detailed real-time visualization—where every item, belt, or robot is tracked on screen from a 2D view. This is coupled with the ability to view monitoring for practically all processes, placing it at the cutting edge of current AI capabilities for handling multimodal data bandwidth and human interface use. 

Bases are commonly developed over several dozens if not hundreds of hours. There is a tremendous amount of temporal information involved in optimizing a base which would certainly test the long-context nature of frontier agents. Sophisticated memory and recall systems would undoubtedly be necessary for an LLM-based agent to succeed in an extended episode playing Factorio. Separately, planning for the future would certainly benefit from time spent reasoning, but this comes at a cost when acting in a real-time environment, hence aligning interplay of System 3 and 4 with a key compute usage trade-off.

Addressing these technical barriers also highlights the importance of multi-agent collaboration: large-scale systems often require multiple agents or human-agent teams working in sync. This necessitates coordination frameworks that facilitate shared state and efficient task delegation. Moreover, real-world complexities like supply-chain delays or hardware breakdowns call for robust decision-making under uncertainty—agents must act swiftly and safely, even with incomplete information. Nevertheless, scaling compute FLOPs and refining AI architectures are likely to improve input-output flow management to the point where agents can handle advanced simulations like \textit{Factorio} in real time (realistically the game only needs to be played at around 5 FPS), without relying on domain-specific observation and action spaces, as was common in earlier superhuman-agent research such as AlphaStar.

\subsection{Modding, Market Interactions, and the Agent-Evaluator Framework}

Factorio’s \textbf{modding ecosystem} is unusually flexible, allowing Lua scripts to fundamentally alter or extend nearly every facet of the simulation. At one end, small “Quality-of-Life” mods streamline actions like inventory management or blueprint deployment—an approach often mirrored in real-world industrial systems where specialized scripts automate repetitive tasks. At the other end, total conversion mods, such as \textit{Space Exploration} \cite{spaceExploration} or \textit{Industrial Revolution 3} \cite{industrialRevolution} introduce entirely new resources, tech trees, and production pipelines. This capacity for extensive re-parameterization means researchers can craft tailored scenarios focusing on, for example, large-scale chemical manufacturing or advanced energy grids. By doing so, Factorio can serve as a robust platform for evaluating AI agents under conditions that closely resemble real-world system engineering challenges.

An especially promising application of this modding framework involves designing \textbf{market pricing} and \textbf{multi-agent} interactions. Factorio already supports multiplayer, and community-created mods showcase how resource trading, diplomatic pacts and emergent economies can drive the game’s complexity \cite{blackMarket, diplomacy}. In a research context, introducing dynamic markets would allow agents to buy and sell resources, negotiate prices, and even form alliances or contracts—key elements of real-world logistics and supply chains. Observing how AI agents adapt to fluctuating market forces and coordinate with others could yield insights into cooperative and competitive strategies, as well as negotiation tactics and resilient system designs.

Beyond market dynamics, Factorio’s modding API also lends itself to the concept of a \textbf{Agent-Evaluator Framework}. In this paradigm, a “evaluator” agent (human or AI) orchestrates scenario constraints, random events, or objectives (Informing system 5 as per the VSM) while the “agent” attempts to build and maintain a functional factory. This setup is well-suited to self play-like reinforcement learning algorithms, where the evaluator can inject perturbations—ranging from supply shortages to power-grid failures—testing the agent’s capacity for adaptive, long-horizon decision-making. The evaluator could also coordinate multiple agents with distinct roles or goals, enabling both collaboration and competition. Such arrangements bring Factorio closer to real-world engineering environments, where teams of engineers and managers must not only design but also continually refactor systems in response to shifting requirements and unforeseen disruptions.

By blending flexible modding, multi-agent mechanics, and the Agent-Evaluator approach, Factorio becomes more than just a factory-building game. It becomes a powerful sandbox for studying how AI agents might operate in large-scale, ever-evolving ecosystems—spanning everything from supply-chain economics to self-directed adaptation and robust error handling. This versatility sets Factorio apart as a uniquely comprehensive testbed for advancing AI-driven system engineering. 

\subsection{Technical Advantages}
Beyond the near-limitless opportunities provided by mods, \textit{Factorio} offers a few key advantages that are worth highlighting. First, as a 2D game, it is far more resource efficient for the complexity of systems that can be built in it as it does not involve costly 3D graphics rendering as would be the case in other titles such as \textit{Satisfactory}. Even despite this major difference, \textit{Factorio} is well-known to be a very well-optimized game in terms of memory usage, as it has been continually refined by its dedicated team since its first public release in 2012. Furthermore, the game is platform-agnostic, running natively on Windows, Mac OS X and Linux, which is rare. It offers a free headless Linux server for supporting well-optimized multiplayer gameplay which would be crucial for human-AI and multi-AI agent experimentation. And as mentioned before, the game has exceptional support for modding, showcased by community mods which completely overhaul the tech tree, environmental mechanics and GUI systems. We believe it is quite feasible to build an API layer for control as an intermediate solution for AI usage similar to Mineflayer \cite{mineflayer2024} (used in the Voyager project \cite{wang2023voyager}). In fact, that could even be a task for an AI agent to perform as part of its introduction to the game.
\section{Alternative Views}

Some critics argue that advancing AI system engineering is premature, given that core capabilities—like consistent reasoning, robust multimodality, and factual grounding—remain underdeveloped. They believe AI should first address these foundational weaknesses before tackling higher-level tasks. Yet proactive, orthogonal research can reveal new performance bottlenecks and drive innovation across modalities. Much as multimodality has progressed alongside unresolved text-based issues, tackling system engineering now can highlight what crucial gaps persist, helping to shape more integrated AI architectures.

Another concern is the risk of entrusting critical infrastructures to automated agents. Misaligned objectives or flawed reasoning could theoretically sabotage energy grids, supply chains, or other vital systems. While these dangers merit attention, the potential benefits—greater efficiency, cost savings, and creative solutions—are substantial. Alignment sits at the core of system engineering, which is rooted in clear requirements, continuous feedback loops, and stakeholder validation. By maintaining transparency and accountability, AI-driven engineering can strike a balance between prudence and progress.

Skeptics may also doubt whether games like \textit{Factorio} adequately reflect real-world complexities, noting they often omit granular physical laws or regulatory constraints. Yet such “unrealistic” environments highlight the essence of system engineering—resource management, strategic planning, and iterative trade-offs in efficiency, adaptability, and cost—far better than static benchmarks and without the noise associated with realistic physics simulations. Skills developed in orchestrating large-scale virtual factories can be paired with domain-specific testing to produce a fuller assessment of AI’s strengths. This integrated approach shows where AI excels (e.g., in macro-level design) and where further refinement is needed before applying these insights to physical-world applications.
\section{Conclusion}
AI agents stand on the verge of a new era where they can systematically design, optimize, and maintain complex systems in ways that rival or surpass human expertise. While LLMs have already showcased impressive capabilities for text generation, the true promise lies in the \textit{agentic paradigm}—with integrated multimodal interfaces, memory, autonomy, and adaptive planning.

We have argued that system engineering represents a high-leverage domain for such agentic AI. Whether the task is orchestrating large-scale software infrastructures or managing logistical networks, adaptability and continuous learning map naturally onto the strengths of a well-trained AI agent. Yet, to properly develop and evaluate these systems, we must look beyond static benchmarks toward open-ended simulations that reflect real-time constraints, multi-agent collaboration, and shifting objectives.

In this regard, \textit{Factorio} emerges as a compelling platform, providing a safe yet rich environment for refining agentic capabilities. Its emphasis on real-time resource management, multi-objective optimization, and large-scale factory layouts makes it a microcosm of industrial-scale challenge. Success in \textit{Factorio} would signal that agents can handle real complexity, track multiple objectives, and adapt in realistic ways.

In conclusion, the evolution from LLM-based chatbots to versatile AI agent that can tackle system engineering marks a logical next step if we hope to solve the grand challenges of our era. By leveraging automation-oriented sandbox simulations like \textit{Factorio}, we can accelerate progress toward AI systems that orchestrate research, design, and operations at scale—fundamentally reshaping how societies function and flourish in the coming decades.

\bibliography{main}
\bibliographystyle{icml2025}

\newpage
\appendix
\section{Visual Introduction to Factorio}
\label{ref:sec_appendix}

This appendix provides a high-level, illustrated overview of key \textit{Factorio} systems, ensuring that newcomers can grasp the fundamental mechanics of extracting resources, setting up production lines, defending against threats, and automating workflows. Each subsection introduces core concepts, from the simplest mining operations to advanced infrastructures like rail networks, circuit logic, and robot-assisted construction.

\subsection{Resource Extraction and Smelting}
%
The foundation of any Factorio factory is consistent raw material throughput. Players begin by placing mining drills on ore patches—such as iron or copper—where the drills extract resources at a steady rate. Ores are usually transported via conveyor belts to nearby smelters, which convert them into plates. A typical early-game setup involves an arrangement of furnaces linked by belts on both the input (ore) and output (finished plates) sides. This workflow underpins the factory’s growth: higher demand for plates necessitates expanding both mining operations and smelting capacity.

\begin{figure}[ht]
    \centering
    \includegraphics[width=\columnwidth]{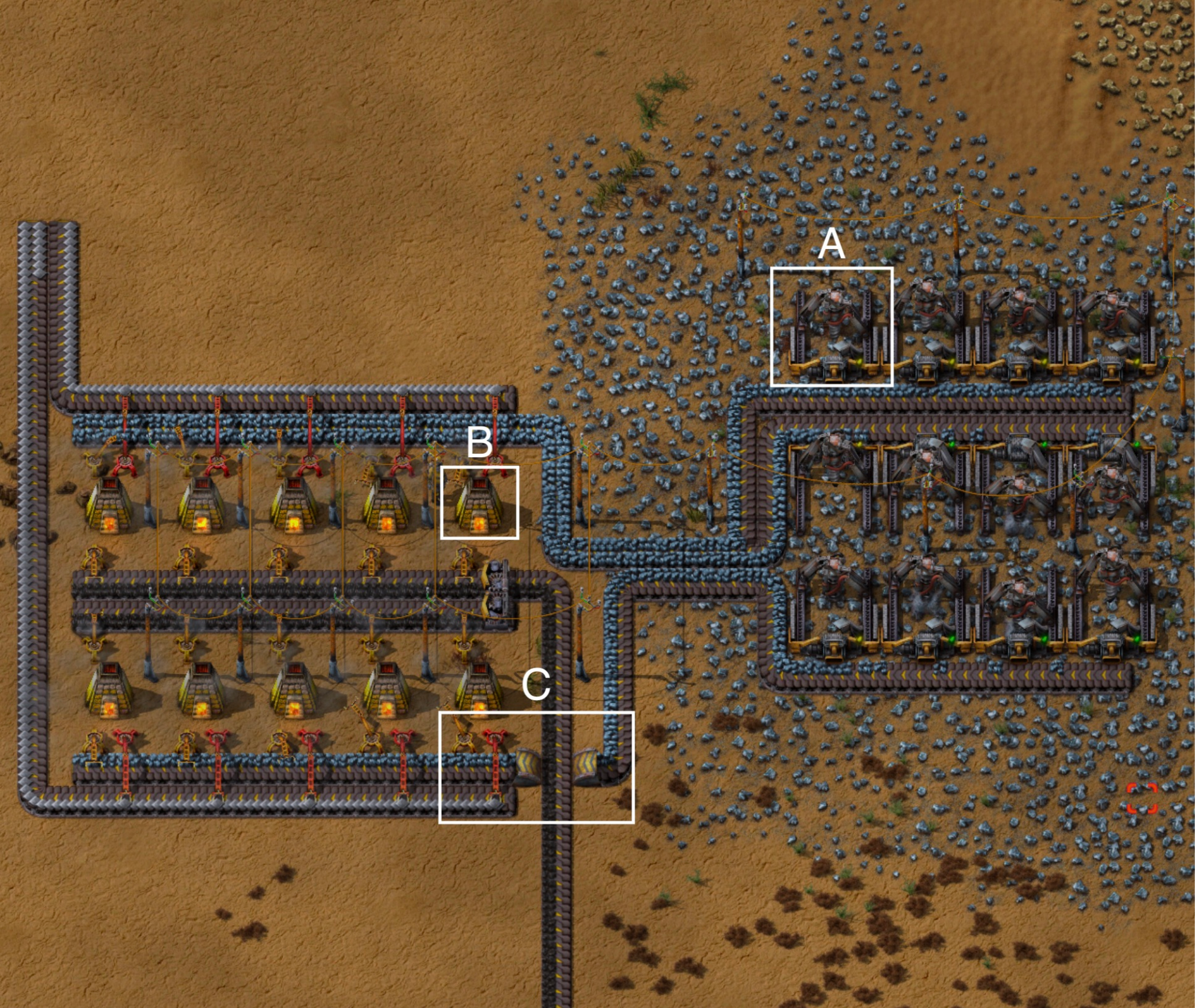} 
    \caption{An example of early-game resource extraction and smelting in Factorio. Box A shows mining drills extracting iron ore, Box B highlights stone furnaces which take ore and fuel and create plates, and Box C highlights belt routing and inserter mechanics. \cite{ironMiningSmelting}}
    \label{fig:iron_smelting}
\end{figure}
We first refer to Figure~\ref{fig:iron_smelting}. The automation process begins with the mining drills in \textbf{Box A}, which extract raw iron ore (blue material) from resource nodes and place it onto conveyor belts for transport downstream. These drills eliminate the need for manual mining, significantly increasing throughput and setting the foundation for automated workflows. In \textbf{Box B}, the raw iron ore is delivered to stone furnaces, where inserters (mechanical arms) feed the ore into the furnaces and remove the resulting iron plates—a critical intermediate resource—onto separate belts. This dual-belt system, fed by both raw iron ore and coal, ensures a continuous and automated smelting process. Finally, in \textbf{Box C}, the belts are routed efficiently using underground segments to avoid intersection conflicts, enabling seamless transport of resources. Yellow inserters deliver iron ore into the furnaces, while red inserters extract the smelted iron plates, which are then routed onward for further processing. This layered system of extraction, smelting, and material routing illustrates the early-game challenges of compact, efficient factory design in Factorio.

\subsection{Automation with Assemblers and Managing Complexity}
\begin{figure}[ht]
    \centering
    \includegraphics[width=0.6\columnwidth]{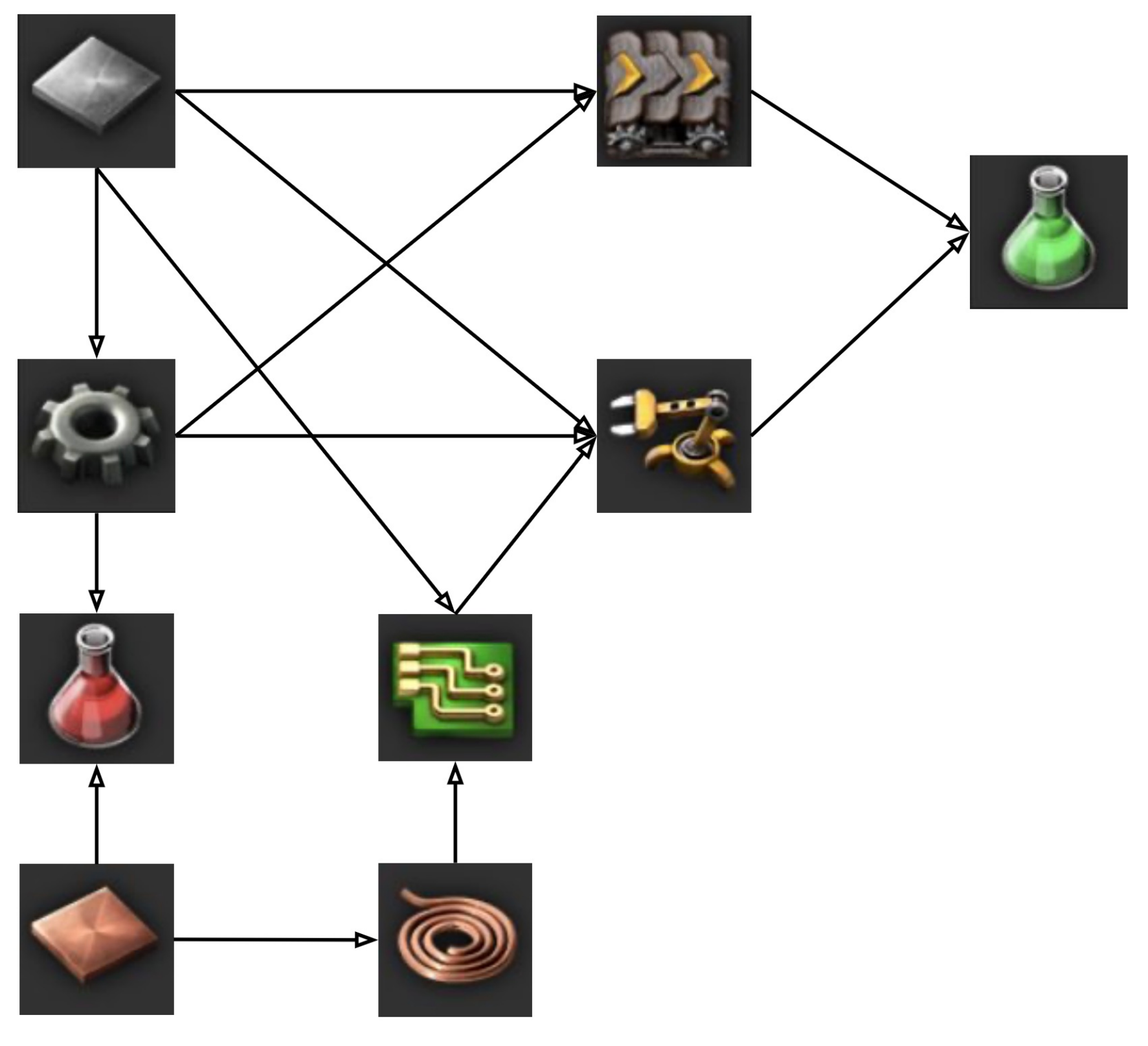}
    \caption{Dependency graph for red and green science packs. Inputs include both raw materials and intermediates, reflecting the growing complexity of production chains.}
    \label{fig:red_green_dependencies}
\end{figure}

\begin{figure}[ht]
    \centering
    \includegraphics[width=0.5\columnwidth]{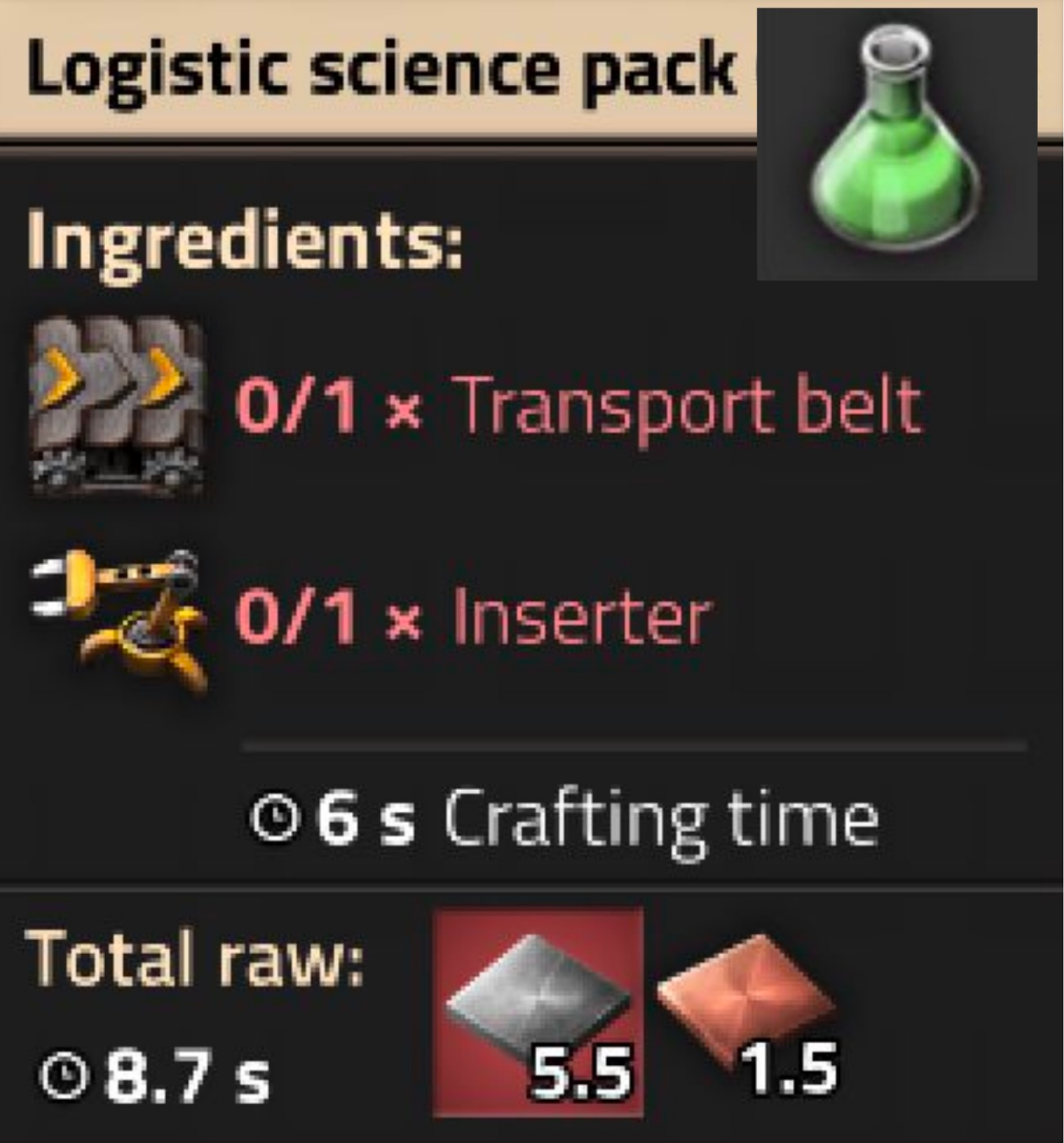}
    \caption{Recipe for logistic (green) science packs. Automating intermediate goods significantly reduces total crafting time from 8.7 seconds (raw) to 6 seconds.}
    \label{fig:green_science_recipe}
\end{figure}

Automation through assemblers is a cornerstone of \textit{Factorio} gameplay, enabling exponential growth in productivity by trading energy and space for vastly higher throughput. The production of \textit{science packs} is central to the objective of unlocking advanced technologies. The earliest science packs are automation (red) and logistic (green) science. The dependencies for crafting these are shown in Figure~\ref{fig:red_green_dependencies}. Taking the example of logistic science, Figure~\ref{fig:green_science_recipe} illustrates that it takes 6s to assemble if the intermediate goods of transport belts and inserters are available. If only the raw materials of iron and copper plates are present then it will take 8.7s since the intermediate goods themselves need to be created. Hence, by automating intermediate goods, the factory can parallelize workflows, ensuring higher efficiency and faster output. 

There are thus many points to consider when designing assembly lines for these finished goods. The throughput of inputs and outputs should be well-matched given the ratios of materials needed in recipes. The demand for a common base resource has to be managed well across different use cases. The factory has to be actively refactored as increased scale means greater space and energy requirements. An example of a compact design which produces both red and green science is shown in Figure~\ref{fig:red_green_assembly}. While it looks efficient, issues may arise when the scale of production needs to increase, since the routing of intermediate goods would be significantly complicated. Efficient layouts must balance immediate needs with scalability, ensuring that adding new production lines or expanding capacity can be achieved without overhauling the entire factory.

\subsection{Science Packs and the Tech Tree}
\begin{figure*}[htb]
    \centering
    \includegraphics[width=0.8\textwidth]{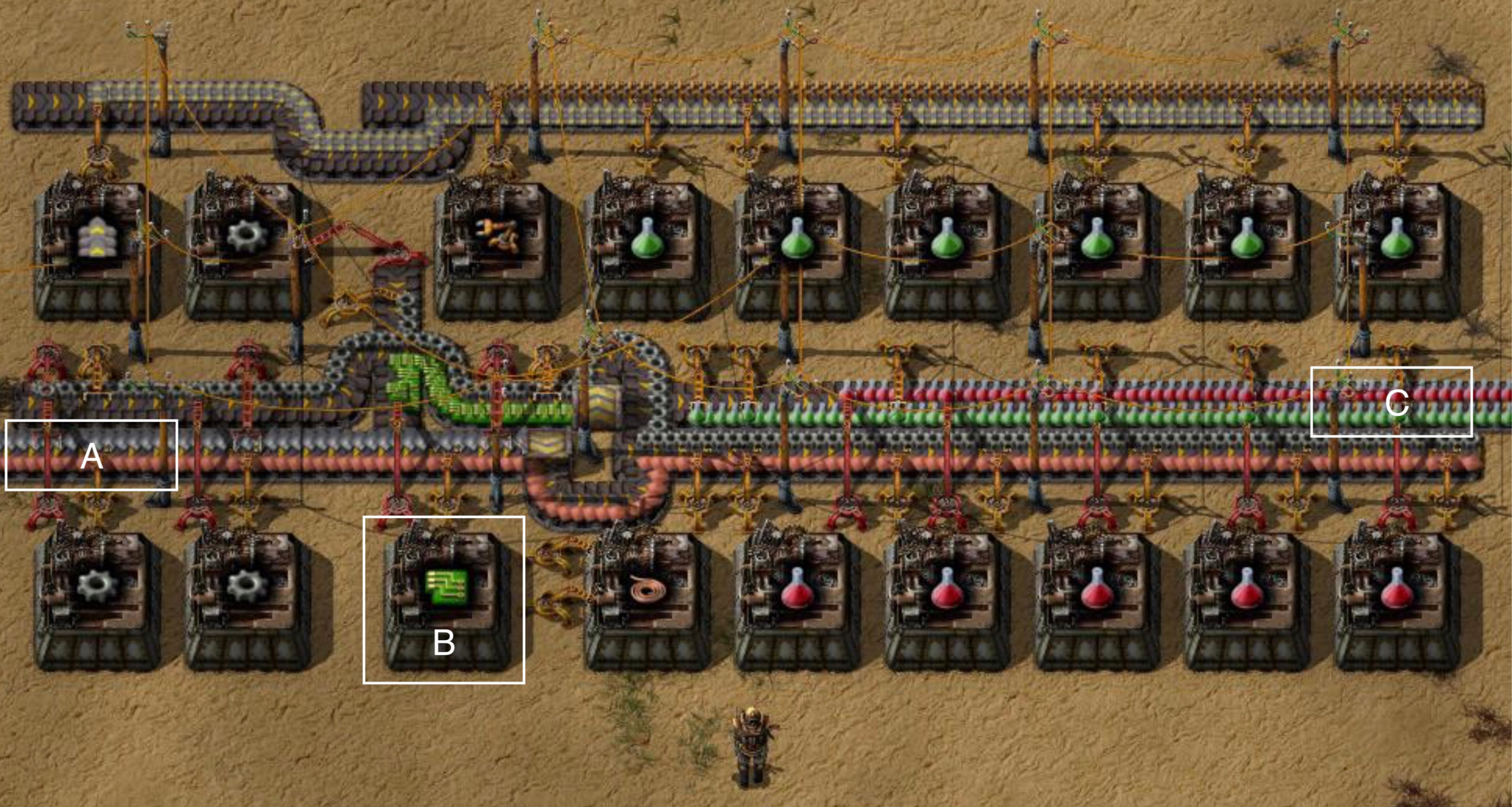}
    \caption{A red and green science production setup. All belts are running from left to right. Iron and copper plates enter on the bottom-most belt (Box A). There are assemblers throughout the line which have certain recipes selected. For example Box B has the assembler responsible for assembling green circuits. The inserters to the right of Box B automatically pull copper wire from that assembler and the yellow inserter above Box B pulls iron plates from the belt. The red inserter above Box B places finished green circuits onto the belt one tile above the belt with iron and copper plates. Similar assembly happens for gears, belts, and inserters. Ultimately, red and green science packs are produced from the intermediate goods and are ready for further use (Box C).}
    \label{fig:red_green_assembly}
\end{figure*}

The science packs produced in Figure~\ref{fig:red_green_assembly} are central to progression in \textit{Factorio}, serving as the primary currency for unlocking new technologies. Science packs are consumed by specialized \textit{laboratory} units, which convert them into research progress. Each research project requires a specific combination and quantity of science packs, introducing dependencies on a wide array of intermediate goods. This makes science packs a natural bottleneck for factory growth, as they represent the culmination of multiple production chains working in harmony.

\begin{figure}[ht]
    \centering
    \includegraphics[width=0.7\columnwidth]{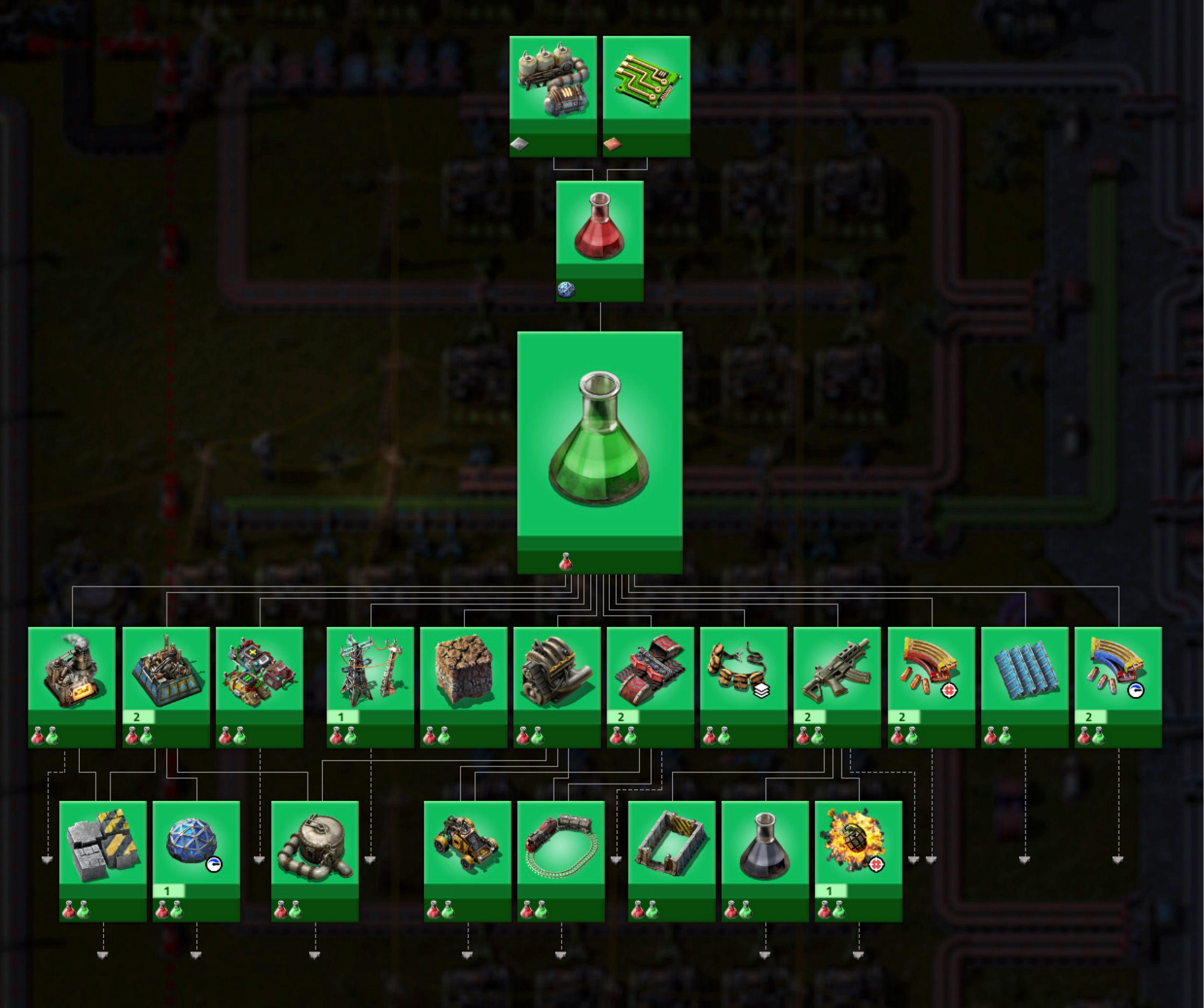}
    \caption{Section of the tech tree which shows which technologies have logistic (green) science as a dependency. These include better transport belts, engines, trains, electric cables, circuit networks and more. Each recipe demands dozens if not hundreds of logistic science packs and so SPM becomes the bottleneck for further growth.}
    \label{fig:green_science_tech}
\end{figure}

This reliance on science packs is why science per minute (SPM) emerges as a critical metric for measuring factory productivity. A high SPM indicates that the factory has sufficient capacity not only to produce the required intermediates efficiently but also to scale them as the tech tree demands become more complex. For example, early-game science packs (red and green) require relatively simple intermediates such as gears, transport belts, and inserters, as shown in Figure~\ref{fig:green_science_tech}. However, as the factory evolves, higher-tier science packs (such as blue or utility science) introduce more advanced recipes involving fluids, electronics, and complex multi-stage production.

The tech tree in Figure~\ref{fig:green_science_tech} highlights this progression, showcasing how early-game technologies provide foundational tools like transport belts and inserters, which are then leveraged to unlock more advanced machinery such as trains and assemblers. This cascading dependency structure requires careful planning to ensure that production systems remain adaptable to increasing demands. The iterative process of unlocking technologies feeds back into the factory itself, enabling further automation and resource optimization. The science system is thus a core gameplay mechanic that ties together automation, logistics, and long-term planning, creating a continuous cycle of technological advancement and production refinement.

\subsection{Power Generation Options}
Factorio offers diverse power solutions that evolve with the factory’s scale, closely mirroring the progression of energy systems in real-world industrial engineering. Early operations rely on steam engines fueled by coal, providing a reliable but resource-intensive solution. Coal mining introduces logistical challenges, requiring consistent supply chains and raising concerns about pollution, which in the game aggravates hostile aliens and causes them to attack the agent's base (see the next subsection). As research progresses, solar panels and accumulators become viable for renewable energy. While solar panels offer clean, sustainable power, they come with limitations tied to diurnal cycles, requiring accumulators to store excess energy for nighttime use. This trade-off between sustainability and infrastructure demands mirrors the challenges of integrating renewables into modern power grids, where storage and energy distribution systems are key bottlenecks.

\begin{figure}[ht]
    \centering
    \includegraphics[width=\columnwidth]{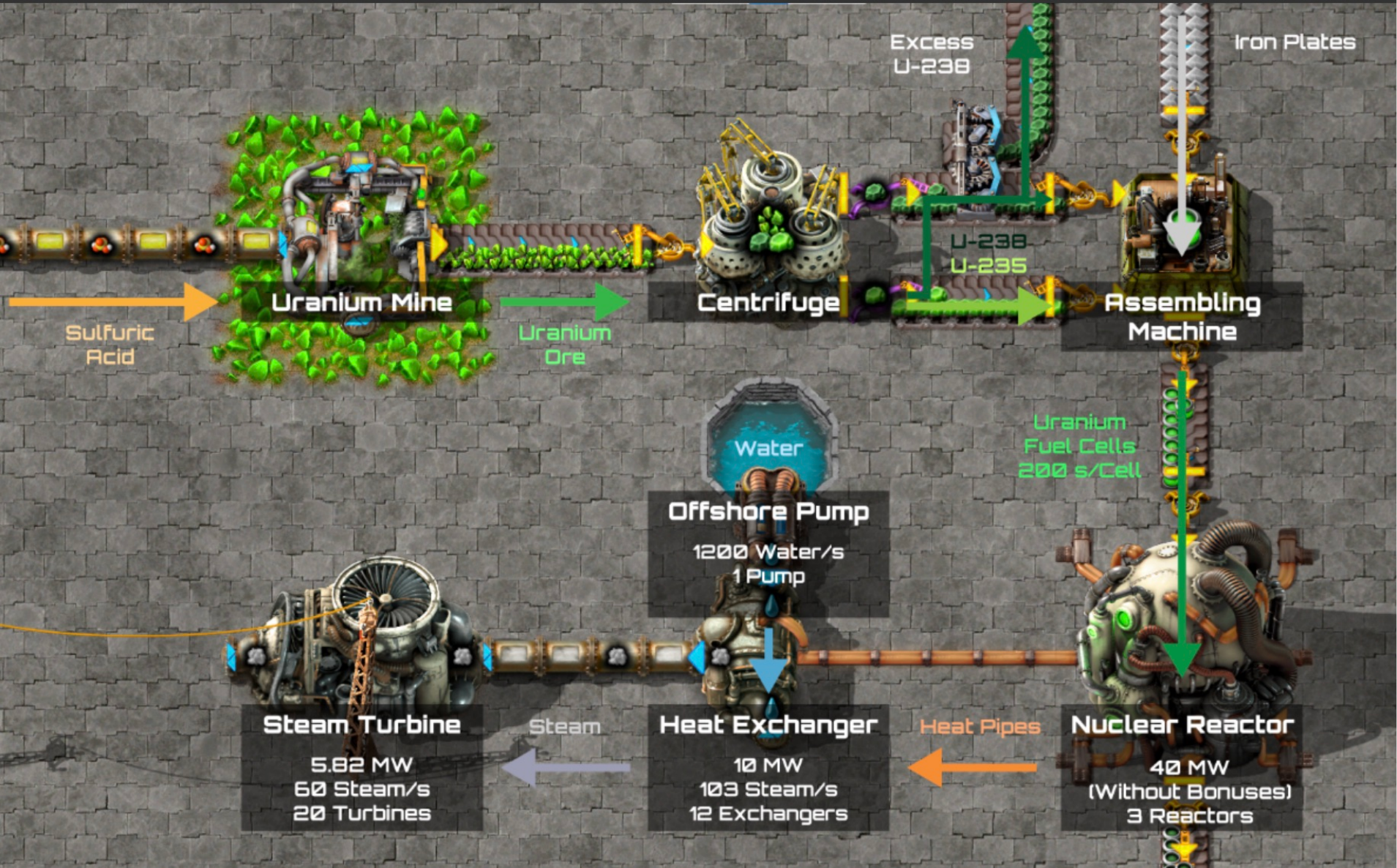}
    \caption{Nuclear power generation is actually quite realistic in \textit{Factorio}. Uranium ore is mined, the vast majority of which (99.3\%) is inert U-238. The more valuable U-235 is needed for energy-intensive applications. There is an enrichment process by which U-238 can be refined to make more U-235 provided some initial quantity of U-235. Then this is utilized in nuclear reactors which can produce steam to power turbines. \cite{nuclear}}
    \label{fig:nuclear_power}
\end{figure}

Nuclear power, a late-game solution, exemplifies high-density energy production but comes with its own complexities. Players must process uranium, manage heat generation, and design safe reactor layouts to avoid catastrophic failures, echoing real-world concerns around nuclear fuel cycles, reactor safety, and waste management. The spatial footprint of energy systems also becomes a critical factor: steam and nuclear setups require compact layouts with high resource input, while sprawling solar farms demand significant land allocation. A reference for nuclear power is shown in Figure~\ref{fig:nuclear_power}

Each energy choice in Factorio presents distinct trade-offs—coal introduces pollution but offers consistency, solar minimizes pollution but requires storage solutions and space, and nuclear delivers immense power but requires advanced materials and precise management. These dynamics force players to weigh efficiency, scalability, and sustainability, capturing the essence of systems engineering in real-world energy infrastructure. By gradually introducing more advanced technologies and requiring players to adapt their power networks, Factorio illustrates the iterative process of scaling energy systems to meet growing demands while addressing environmental and logistical constraints.

\subsection{Biters and Defense}
\begin{figure}[ht]
    \centering
    \includegraphics[width=\columnwidth]{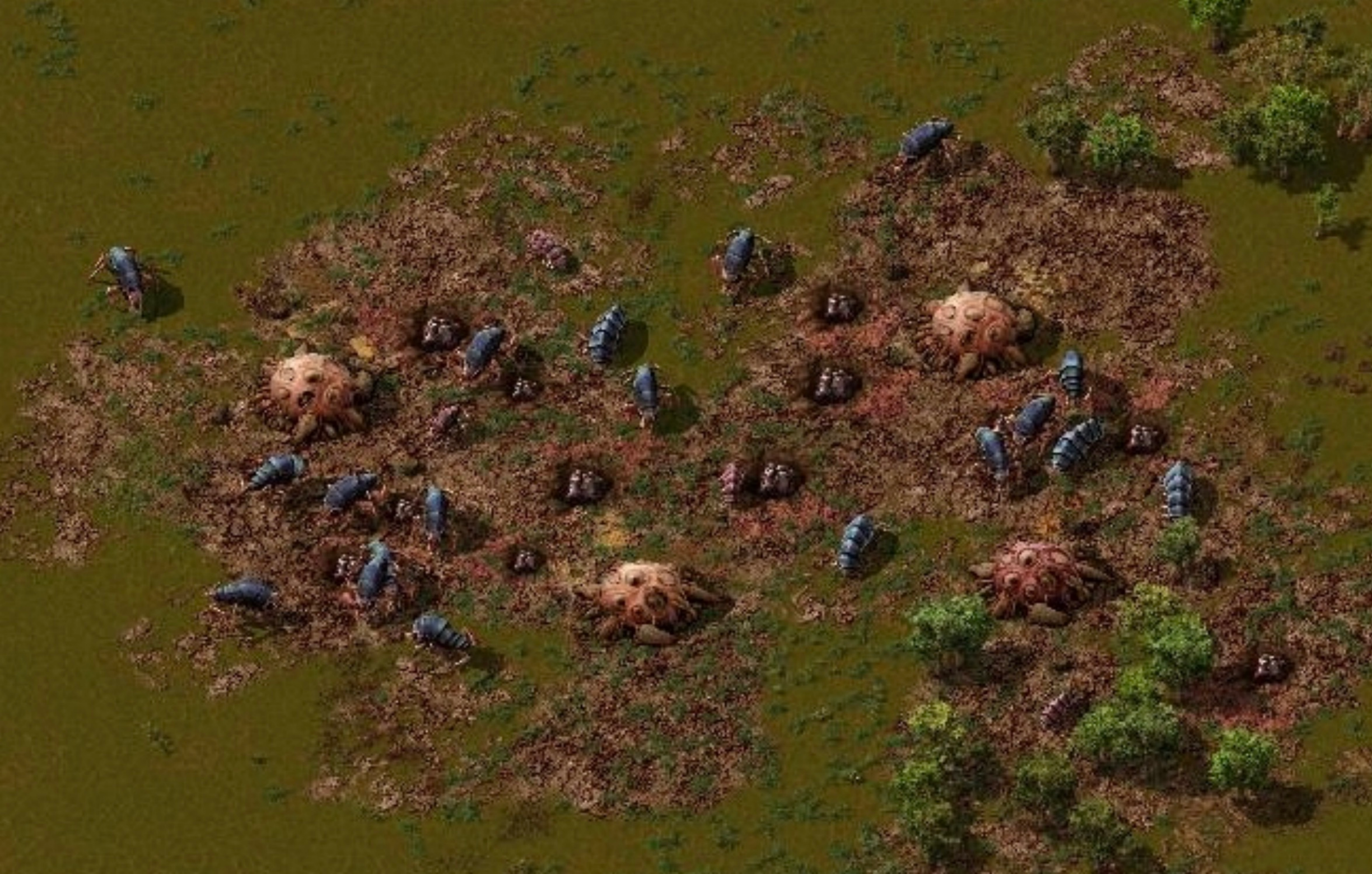}
    \caption{Biters are alien residents of the planet where the agent has crash landed. They are docile initially but become aggravated by air pollution from the factory's hydrocarbon-powered operations. Thinking about biters is thus a core trade-off of expanding systems in Factorio.}
    \label{fig:biter_colony}
\end{figure}

As factories grow and produce pollution, the indigenous alien lifeforms—commonly called Biters—become increasingly hostile, posing a persistent threat to factory operations. Pollution emitted by the factory spreads across the map, and once it reaches a Biter colony, such as the one depicted in Figure~\ref{fig:biter_colony}, it triggers aggressive behavior. Biters begin spawning in waves to attack the factory, targeting structures and resources critical to production. This introduces a dynamic tension between industrial expansion and the need to secure valuable infrastructure, reflecting real-world trade-offs in industrial development where growth often necessitates heightened security measures.

\begin{figure}[ht]
    \centering
    \includegraphics[width=\columnwidth]{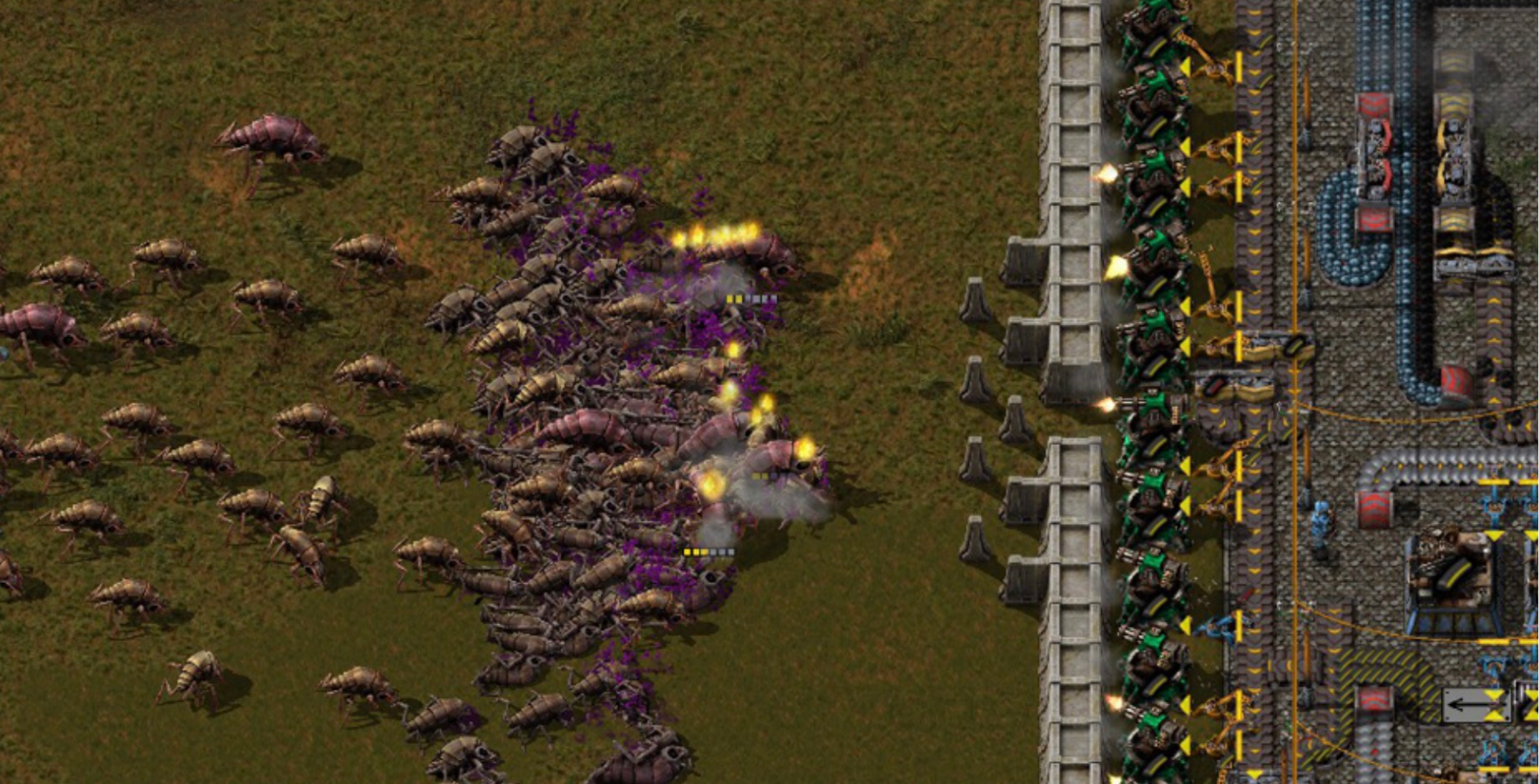}
    \caption{Defense against Biters is essentially a resource sink in \textit{Factorio}. Settings and mods can be used to dramatically change the difficulty associated with defending bases from Biters.}
    \label{fig:biter_defense}
\end{figure}

Early defenses rely on a combination of walls and gun turrets, as seen in Figure~\ref{fig:biter_defense}, where turrets gun down an approaching wave of Biters at a fortified perimeter. Gun turrets provide reliable protection during the early stages but depend on a steady supply of ammunition, which itself requires dedicated production lines. As the factory evolves, more advanced defensive structures like flamethrower turrets, laser turrets, and artillery become available. Flamethrowers are particularly effective for handling large swarms due to their area-of-effect damage, while laser turrets require no ammunition but demand significant power, introducing another layer of logistical complexity. Artillery, a late-game option, allows players to strike Biter nests at long range, proactively reducing the threat level.

Strategically fortifying perimeters and clearing nearby Biter nests becomes essential as pollution spreads farther and factory operations grow in scale. Defensive layouts must balance resource efficiency with resilience, ensuring that critical areas are well-protected without overextending the factory’s capacity to supply power, ammunition, or repairs. Additionally, players must consider choke points, turret placement, and overlapping fields of fire to maximize defensive effectiveness.

\subsection{Complex Belts and Main Factory Layouts}
%
%
Conveyor belts are the arteries of a \textit{Factorio} base, transporting materials between production stages with speed and efficiency. While straightforward in the early game, managing belts becomes increasingly complex as factories grow. Scaling introduces challenges such as belt congestion, balancing input and output ratios, and ensuring that each production branch receives the right materials without overloading the system. Designing efficient layouts to manage this complexity is critical for avoiding “spaghetti”—a term used by the community to describe tangled, chaotic belt arrangements that hinder scalability and troubleshooting.

\begin{figure}[ht]
    \centering
    \includegraphics[width=\columnwidth]{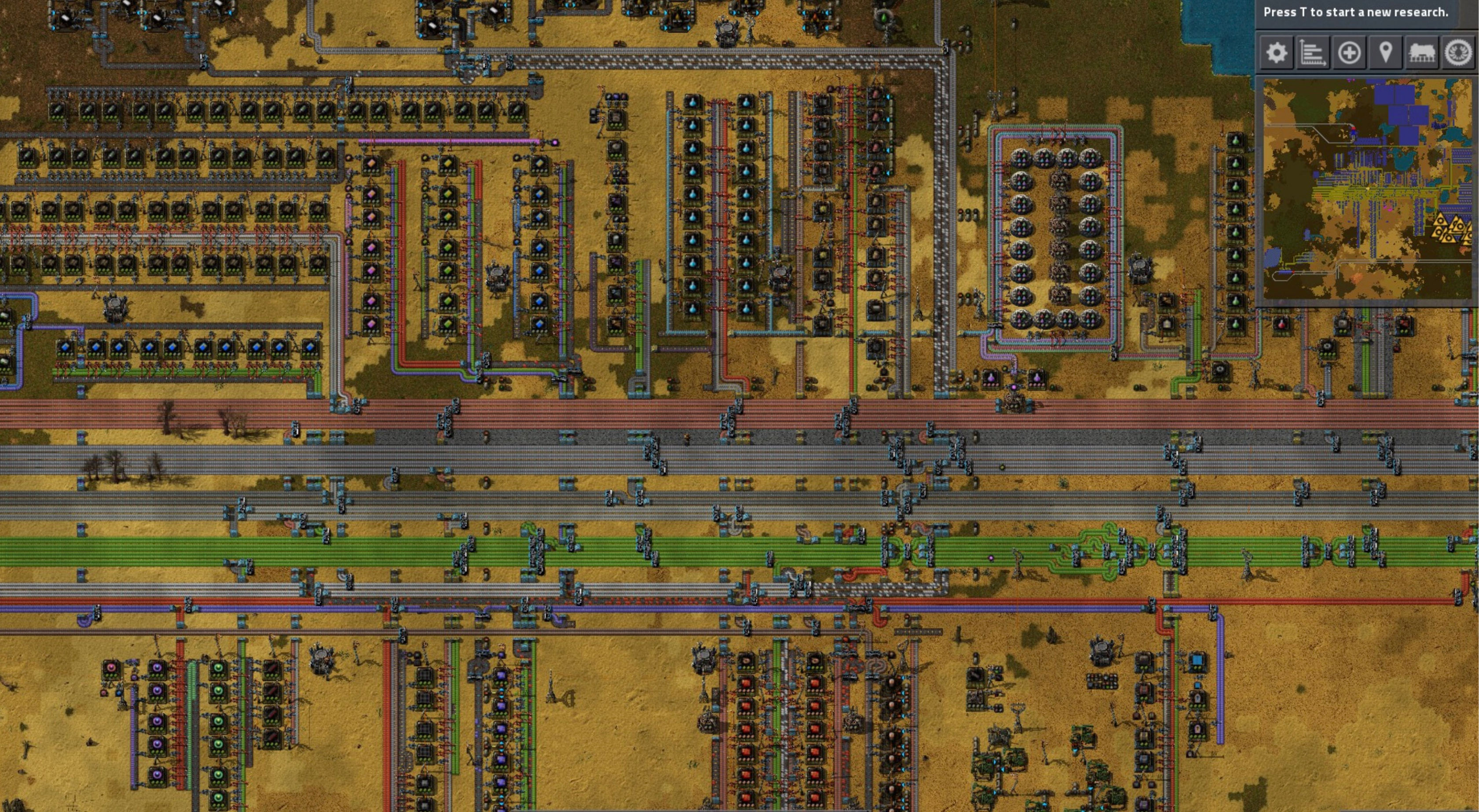}
    \caption{The main bus design is a common choice for mid-game scaling. Branches for individual component assembly fork off the main bus using belt splitters and underground belts. \cite{mainBus}}
    \label{fig:main_bus}
\end{figure}

One popular solution to these challenges is the \textbf{main bus} design, as shown in Figure~\ref{fig:main_bus}. A main bus consists of a centralized set of parallel belts carrying essential resources like iron plates, copper plates, gears, and circuits. Branches extend from the main bus to feed production lines, ensuring that critical resources are readily available across the factory. This design prioritizes simplicity and organization, making it easier to scale production by adding new branches or extending the bus itself. However, maintaining a main bus requires careful planning to prevent bottlenecks and to allocate space for future resource additions. Players must also ensure that belts remain balanced to avoid starving downstream branches of materials.

\begin{figure}[ht]
    \centering
    \includegraphics[width=\columnwidth]{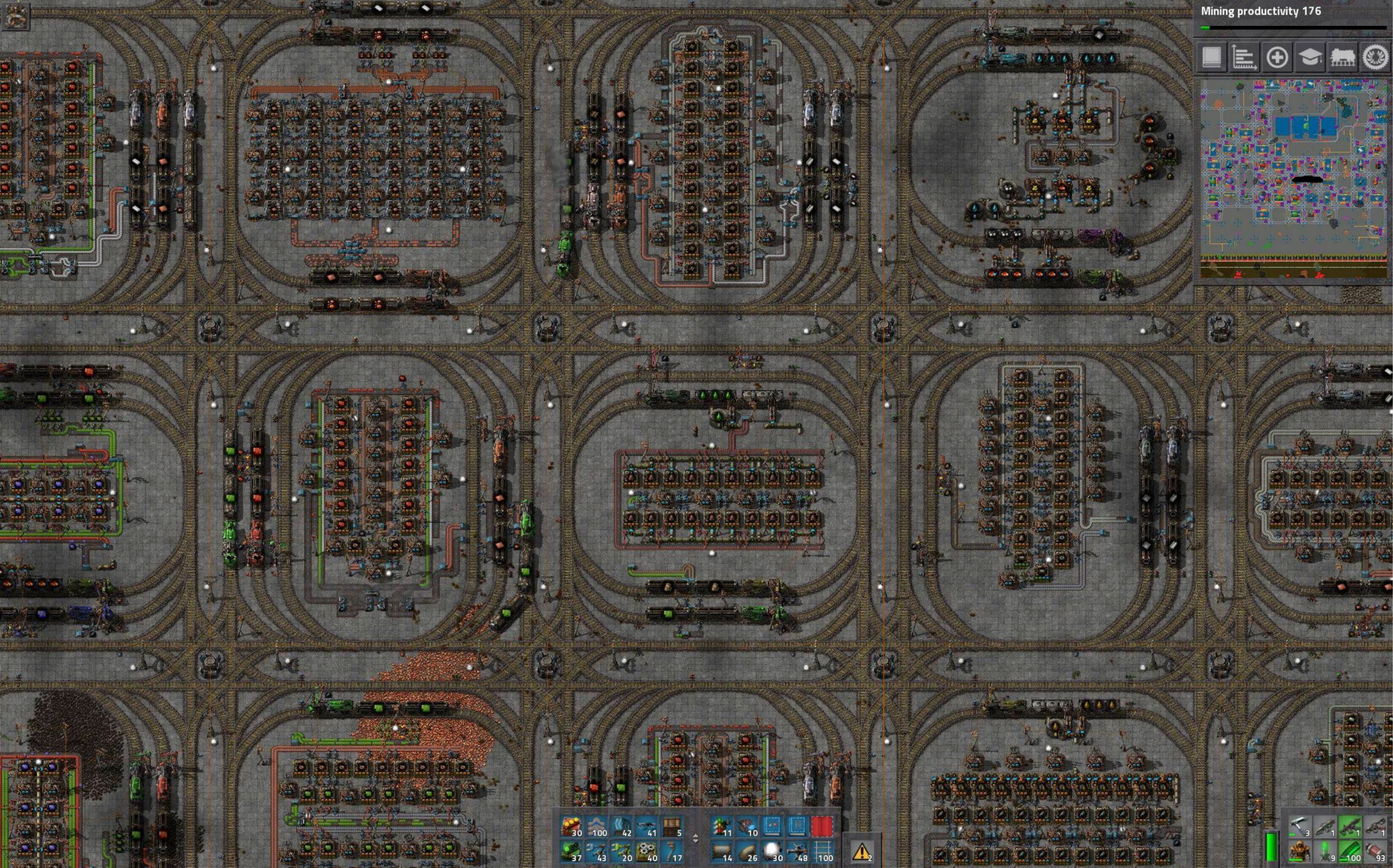}
    \caption{The city block design is ideal for late-game mega-base building. Modular base sections are linked using rail networks for loading and unloading of items. \cite{cityBlock}}
    \label{fig:city_block}
\end{figure}

An alternative to the main bus approach is the \textbf{city block} design, illustrated in Figure~\ref{fig:city_block}. In this modular approach, the factory is divided into distinct “blocks,” each dedicated to a specific function, such as smelting, circuit production, or science pack assembly. These blocks are connected by train networks, allowing resources to be transported efficiently between distant sections of the factory. The city block layout offers excellent scalability, as additional blocks can be added without disrupting existing workflows. It also improves manageability, as each block operates semi-independently, reducing the risk of widespread factory failures due to localized issues.

Both layouts demonstrate distinct trade-offs. The main bus design excels in compactness and simplicity, making it ideal for medium-sized factories, but it can become unwieldy as the number of resources grows in the late game. City block layouts, while more complex to set up initially, provide unmatched flexibility and extensibility, especially when managing large-scale operations with diverse production needs. Figures~\ref{fig:main_bus} and~\ref{fig:city_block} highlight the strengths of these designs, showcasing how thoughtful belt organization and transportation planning are essential for managing complexity and ensuring smooth factory operation as production demands increase.

\subsection{Rail Networks}

At mid to late stages of \textit{Factorio}, trains become a critical component of resource logistics, allowing raw materials and finished goods to be transported across vast distances. Tracks are laid out on a tile-based map, with stations configured for specific tasks such as ore pickups and deliveries to smelting or assembly sites. Trains enable players to overcome the limitations of conveyor belts, which can become cumbersome and inefficient for long-range transport, providing a scalable solution that supports factory growth.

\begin{figure}[ht]
    \centering
    \includegraphics[width=\columnwidth]{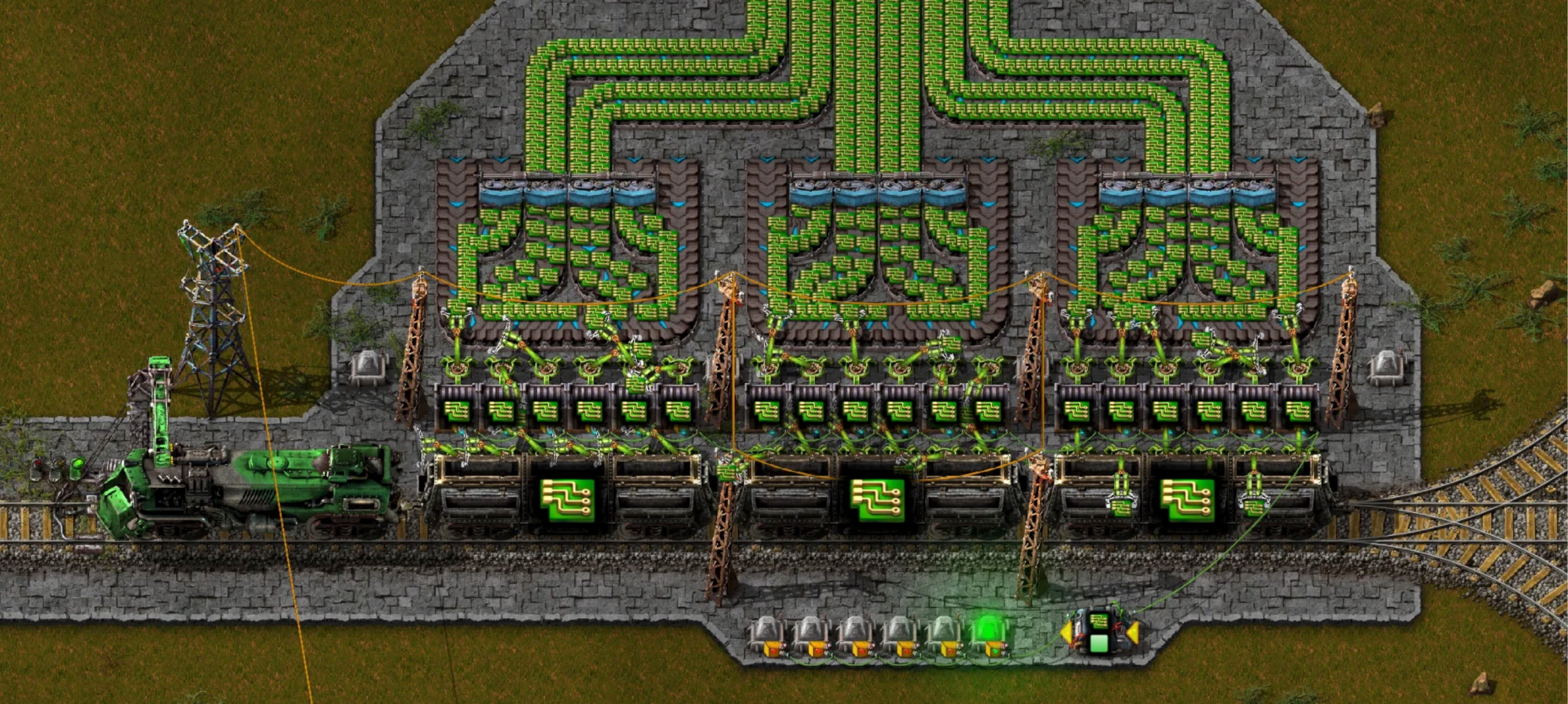}
    \caption{Trains can move large quantities of resources long distances much faster than belts while reusing the same underlying infrastructure, making them crucial for any scalable build. \cite{trainUnloading}}
    \label{fig:_train_loading}
\end{figure}

Designing a robust railway system requires careful planning and mastery of key mechanics. Figure~\ref{fig:_train_loading} shows a typical train loading setup, where numerous inserters work in parallel to load ore into cargo carriages quickly. Efficient loading and unloading are essential to minimize train idle times and ensure smooth throughput. Multiple stations can be linked along a track network, with each station named and assigned schedules dictating when trains should arrive and depart, further streamlining the flow of resources between locations.

\begin{figure}[ht]
    \centering
    \includegraphics[width=\columnwidth]{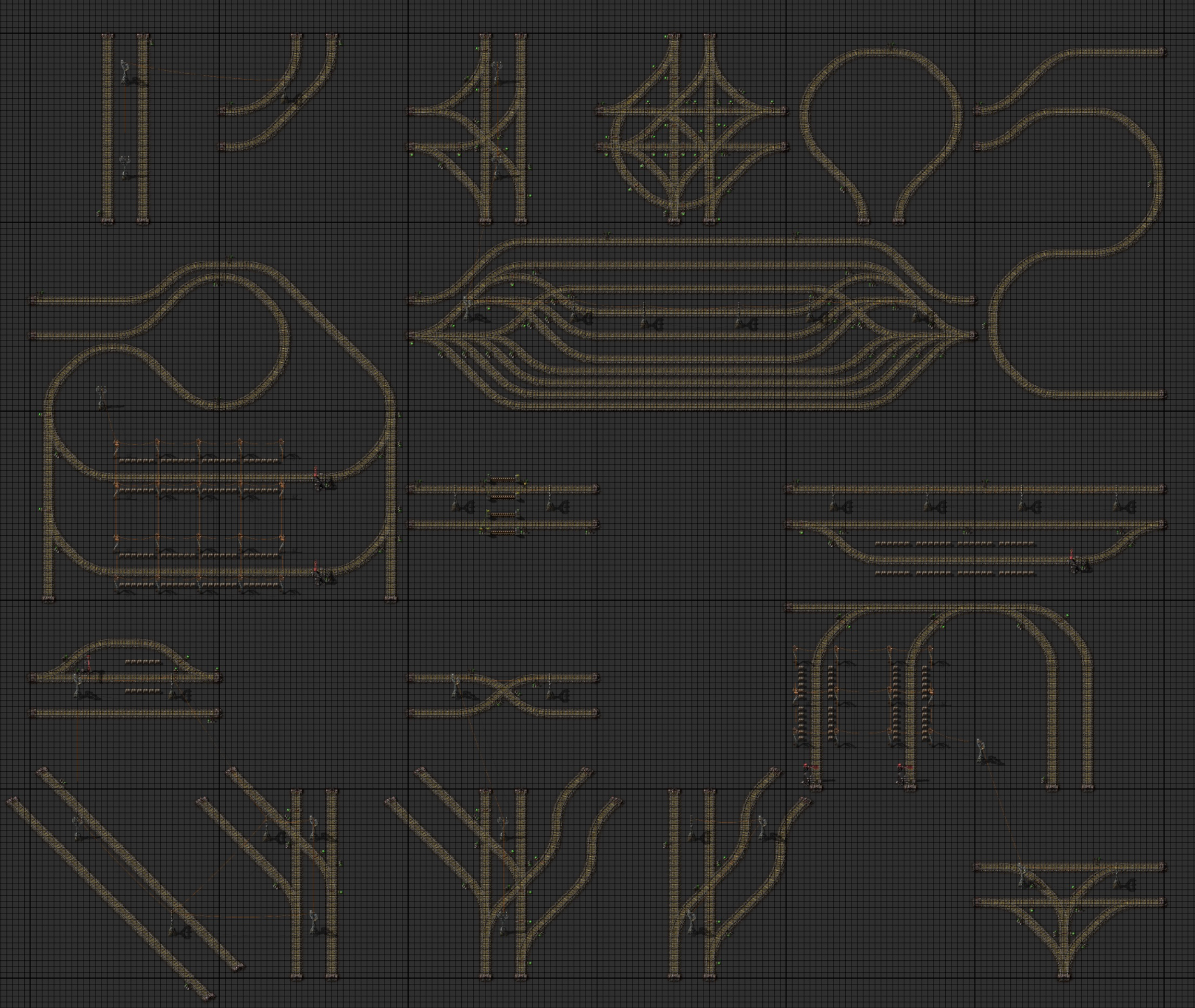}
    \caption{The variety and customization associated with building rail networks is vast in Factorio. \cite{trainPatterns}}
    \label{fig:_train_intersections}
\end{figure}

Track layouts, particularly intersections and junctions, are another crucial element. Figure~\ref{fig:_train_intersections} highlights several blueprint designs for rail intersections, showcasing patterns optimized for traffic flow and collision avoidance. Proper signaling is necessary to manage multiple trains on shared tracks, with block signals and chain signals controlling which sections of track are reserved for individual trains. Complex networks can handle dozens of trains simultaneously, but poor design or inadequate signaling can lead to congestion or catastrophic collisions, disrupting the factory’s supply chains.

\begin{figure}[ht]
    \centering
    \includegraphics[width=\columnwidth]{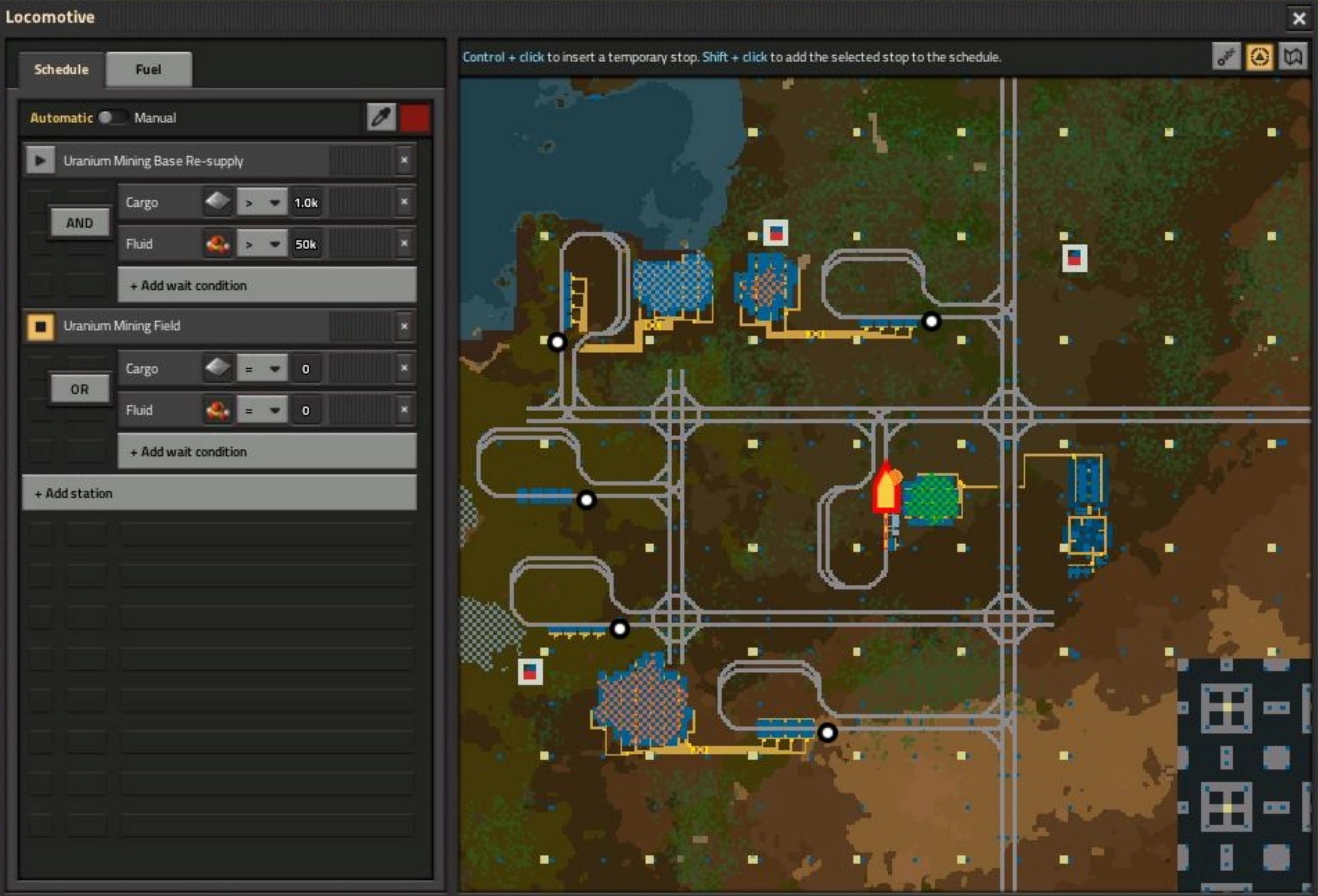}
    \caption{Factorio gives players the ability to observe and orchestrate train networks with high customization \cite{trainGuide}}
    \label{fig:_train_monitor}
\end{figure}

The train management system extends beyond physical tracks, as shown in Figure~\ref{fig:_train_monitor}, which displays the GUI for monitoring train activity. This interface allows players to track the status of all trains in the network, observe their current locations, and adjust schedules or routes as needed. The train monitor is an invaluable tool for diagnosing delays, optimizing routes, and ensuring that all resource flows remain balanced.

A well-designed railway system is not just a means of transport but a backbone for factory expansion, allowing new outposts and production sites to be integrated seamlessly into the larger network. By balancing efficient loading, modular track designs, and robust train management, players can scale their factories to unprecedented levels while maintaining resource flow and minimizing logistical bottlenecks.



\begin{figure}[ht]
    \centering
    \includegraphics[width=\columnwidth]{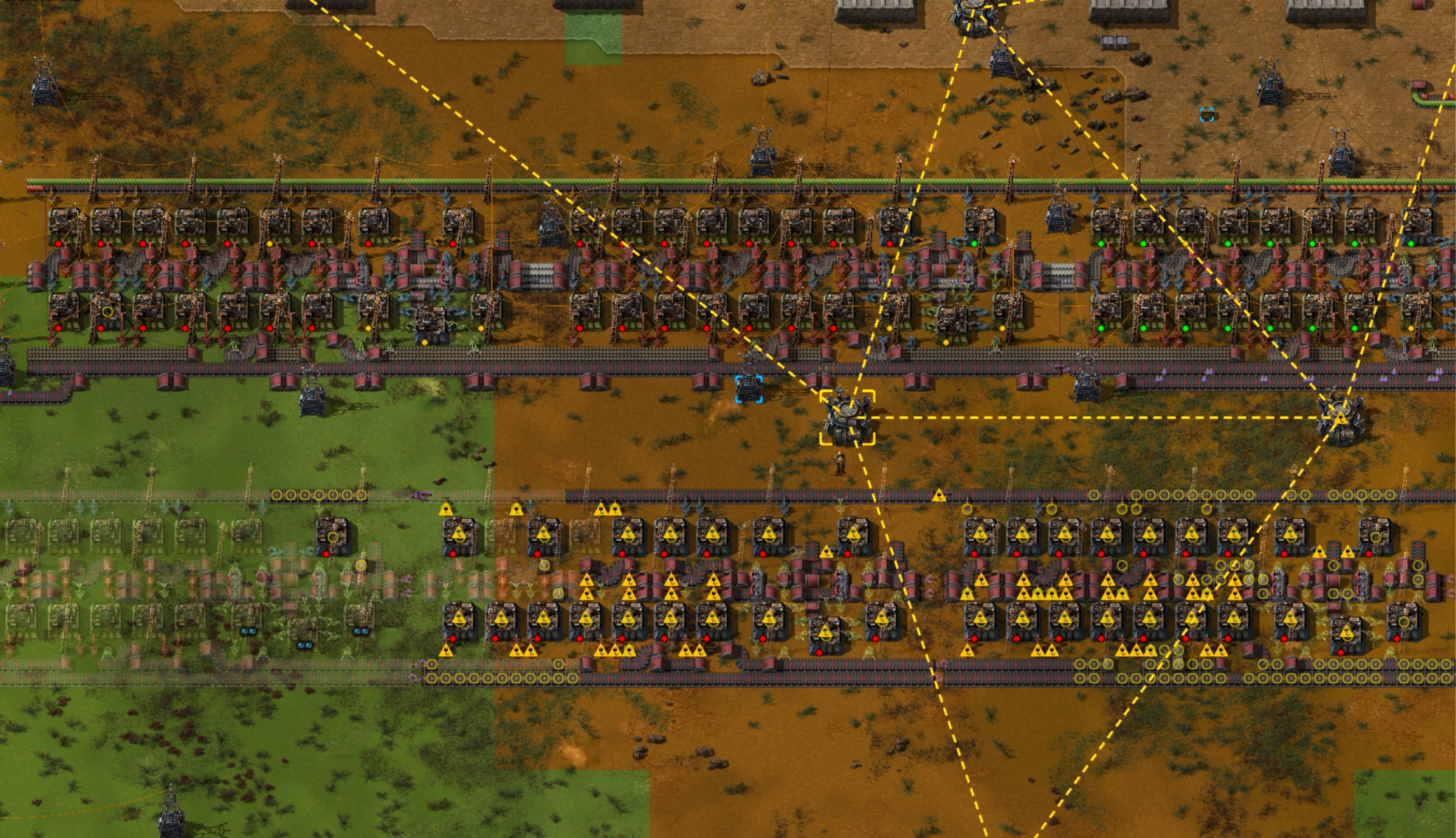}
    \caption{Construction robots automate the placement of arbitrarily complex player-made blueprints. Here the blueprint has been partially constructed by robots and needs to be completed and connected to a source of power. \cite{constructionBots}}
    \label{fig:constructionBots}
\end{figure}

\subsection{Inventory, Blueprints, and Construction Robots}
Players interact with a comprehensive inventory GUI that tracks personal items, crafting queues, and equipment. This interface underpins many of the high-level systems within Factorio, ensuring that even the most complex production chains remain manageable. When testing these systems—particularly in large-scale or late-game scenarios—developers and players alike must verify that inventory updates, crafting queues, and personal equipment management work seamlessly without bottlenecking progress. Such testing is crucial because any inefficiency or bug in inventory handling can cascade throughout a massive base, undermining the player’s ability to grow their automation network.

A prime example of Factorio’s advanced systems is the \textbf{blueprint} feature, which allows users to save layouts ranging from simple assembler setups to sprawling smelter arrays. As shown in Figure \ref{fig:constructionBots}, pasting a blueprint summons construction robots to automatically assemble buildings and belts, provided that the necessary items are available and that the structures remain within the logistic network’s coverage (the robot hub range is visible in the center of the screenshot). High-level system testing involves confirming that these blueprint placements work at scale: robots must reliably build, repair, and upgrade components in the correct order and handle resource shortages gracefully. If the blueprint system or robot AI malfunctions, it can cause partial constructions or idle bots, quickly eroding the advantages of automation and frustrating the player.

\begin{figure}[ht]
    \centering
    \includegraphics[width=\columnwidth]{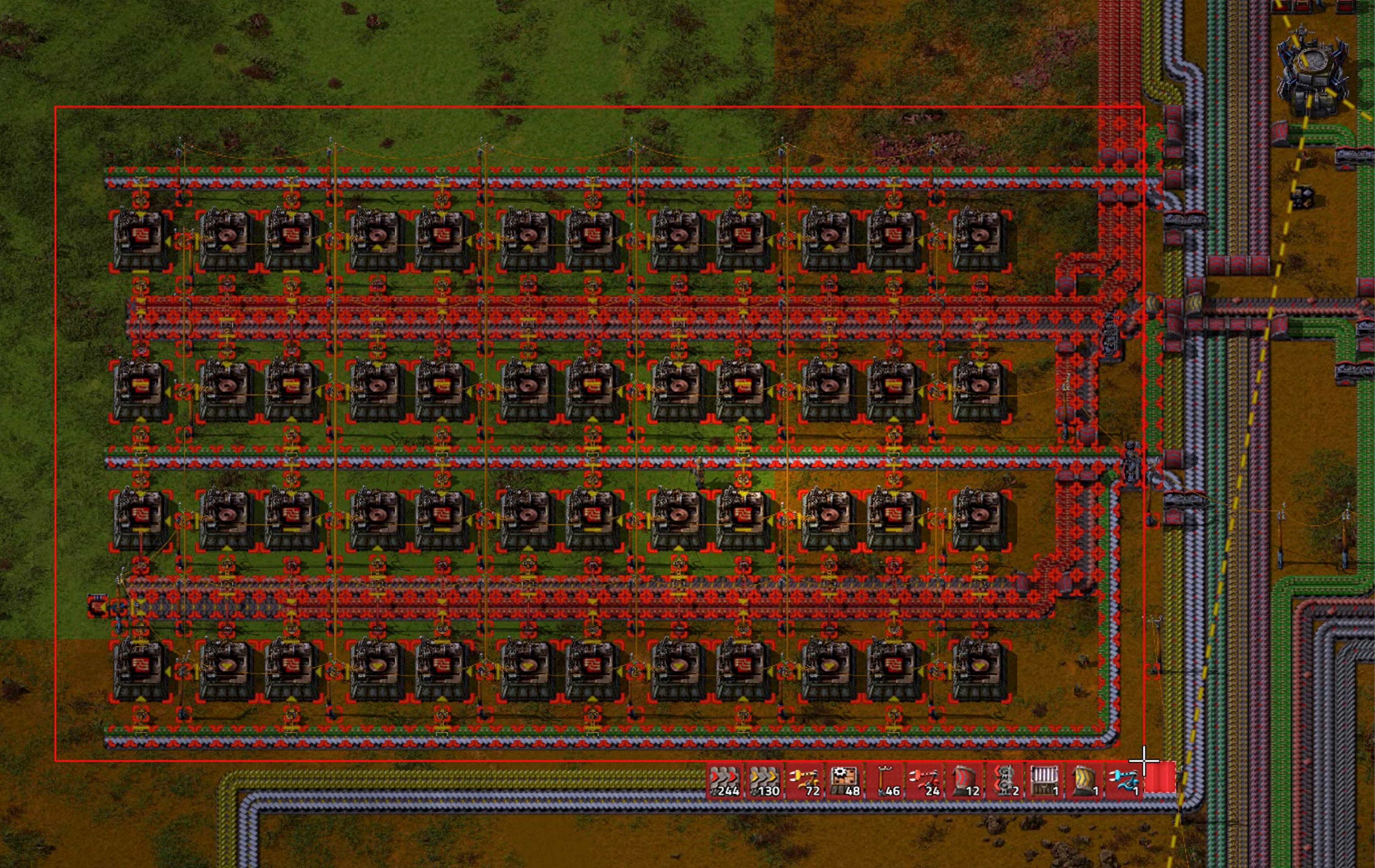}
    \caption{Robots can also be used to efficiently clear out a factory and reclaim the resources. Here the section of the factory has been marked for clearance and robots will swarm it when the player finalizes the selection. \cite{destructionBots}}
    \label{fig:destructionBots}
\end{figure}

Moreover, these same construction robots facilitate large-scale deconstruction, an equally vital aspect of advanced base management. Figure \ref{fig:destructionBots} illustrates the user highlighting a section of the factory for removal—once marked, robots swarm to dismantle it, returning valuable materials to the appropriate storage points. Rigorous testing here ensures that no mismatches occur in item retrieval, that robots can safely access all structures slated for removal, and that the logistic system manages reclaimed items without jams. Essentially, blueprints and bots automate both production and the \textit{creation of production}, making the entire game experience highly recursive and reliant on flawlessly functioning high-level systems. Verifying these features in complex, large-scale conditions is critical for preserving Factorio’s hallmark sense of continual, smoothly scaling automation.

\end{document}